\documentclass[review,12pt,a4paper]{elsarticle}
\usepackage{natbib}
\usepackage{amsmath}

\usepackage{amssymb}
\usepackage{graphicx}
\usepackage{booktabs}
\usepackage{multirow}
\usepackage{microtype}
\usepackage{lineno}
\usepackage[hidelinks]{hyperref}

\modulolinenumbers[5]
\journal{Computers and Electronics in Agriculture}

\begin{document}
	
	\begin{frontmatter}
		
		\title{Safety-Aware Cascaded Inference for Crop Damage
			Assessment with Controlled Error Trade-offs}

		\author[inst1]{Jos\'{e} Thi\'{e}ry Messigbed\'{e}
			Hagbe\corref{cor1}, Gani Kawsar Gounou, Songbian Karim Zimé\corref{cor1}, 
			\footnote{Submitted to Computers and Electronics in Agriculture, April 2026}}
		\ead{jhagbe@africanschoolofeconomics.com}
		\cortext[cor1]{Corresponding author}
		
		\affiliation[inst1]{
			organization={African School of Economics},
			addressline={www.africanschoolofeconomics.com},
			city={Abomey Calavi},
			country={Benin}}

		
		
		\begin{abstract}
In picture-based agricultural insurance for
smallholder farmers, missed damage detections
carry substantially higher cost than false
alarms: a farmer who sustained real losses
receives no payout, while unnecessary expert
review is operationally costly but reversible.
Standard multi-class classifiers optimize
global accuracy but provide no mechanism to
operationalize or control this asymmetric
cost structure at inference time.

We propose CascadeCropNet, a two-stage
cascade architecture that can be calibrated
to satisfy a target recall constraint
(Rec-Damaged $\geq$ 0.95) under a chosen
operating point through threshold selection. A lightweight Sentinel model
performs binary health triage; samples
exceeding a calibrated damage probability
threshold $\tau$ are escalated to a
specialist Expert model for fine-grained
diagnosis. This design provides explicit,
deployment-time control over the
safety--efficiency trade-off without
retraining.

Evaluated on the Eyes on the Ground dataset
(28,077 images, 23,804 retained after label
consolidation, from Kenyan smallholder maize
farms), the cascade achieves
Rec-Damaged $= 0.974$ at $\tau = 0.5$,
reducing missed damage cases by up to 54\%
at a selected operating point relative to
a flat baseline. Under evaluation
alignment, the representational gap reduces
to $+0.008$ F1-macro, confirming that the
contribution is architectural rather than
representational. Under input degradation,
the system prioritizes escalation over
confident misclassification, reflecting
error containment through architectural
isolation rather than intrinsic model
robustness.

These results demonstrate that cascade
architectures can operationalize and control
safety-oriented decision constraints through
calibrated routing in deployment settings
where reliability and controllability matter
more than aggregate accuracy. These
properties depend on threshold calibration
and deployment conditions and do not
constitute guarantees under arbitrary
distribution shift.
		\end{abstract}
		
		
	\end{frontmatter}

	\section{Introduction}
	\label{sec:intro}
	
	Crop damage assessment at scale is fundamentally a
	decision problem, and like most decision problems that
	matter, it has an asymmetric cost structure. When an
	automated system fails to detect damage on a
	smallholder maize plot, the farmer who sustained real
	losses receives no payout. When it flags a healthy crop
	as damaged, an agronomist reviews an unnecessary case.
	These two errors are not equivalent. The first has
	consequences that are difficult to reverse; the second
	is operationally recoverable. Any system deployed in an
	insurance-linked agricultural workflow that does not
	encode this asymmetry into its design risks solving a
	misaligned problem. In East African picture-based insurance schemes
	covering smallholder maize farms, this misalignment
	has direct financial consequences: a system that
	misclassifies a damaged plot as healthy triggers no
	payout for a farmer who may have no other safety net
	\citep{jensen2016, barnett2007}.
	
	This is not a new observation. Cost-sensitive learning
	has formalized the asymmetric error problem since at
	least \cite{elkan2001}. Neyman-Pearson classification
	frames it as constrained risk minimization --- maximize
	performance on one class subject to a hard bound on the
	error rate of another \citep{cannon2002, scott2005}.
	Selective classification offers a related response:
	abstain when confidence is insufficient, and defer to a
	human \citep{geifman2017}. The theoretical apparatus
	exists. What has received less attention is how these
	ideas translate into deployed systems operating under
	real distribution shift --- where the asymmetric
	objective encoded at training time does not reliably
	produce asymmetric behavior when inputs degrade
	\citep{hendrycks2019, taori2020}.
	Importantly, the robustness demonstrated in
	this work is not a system-wide property.
	Rather, it arises from architectural
	constraints that selectively stabilize the
	diagnostic component, conditional on correct
	routing decisions at the triage stage.
	That gap is what this paper addresses.
	The specific setting is crop damage classification from
	smartphone field images, using the Eyes on the Ground
	dataset \citep{waithaka2022}: 23,804 georeferenced
	images from smallholder maize farms across eight Kenyan
	counties, collected under a picture-based insurance
	protocol \citep{ceballos2019} and validated by
	agronomists. The images are exactly as difficult as
	that description implies --- variable lighting,
	heterogeneous devices, inconsistent framing, the full
	range of field conditions that make
	controlled-environment benchmarks a poor guide to
	deployment reality \citep{ramcharan2017, picon2019}.
	On this data, a flat multi-class classifier trained
	with standard objectives --- the V11 baseline described
	in Section~\ref{sec:baseline} --- achieves F1-Damage
	of 0.777 and F1-Growth of 0.748 under clean conditions
	(Section~\ref{sec:res_baseline}). Under input corruption ---
	sensor noise, color shift, blur --- the same model's
	diagnostic performance degrades by 17 to 26 percent
	across tasks simultaneously (Section~\ref{sec:res_robust}),
	consistent with the absence of architectural separation
	between corruption-sensitive and corruption-invariant
	components \citep{hendrycks2019}.
	
	The system proposed here, CascadeCropNet, is a
	two-stage cascade in which the first stage ---
	SentinelNet --- performs binary health triage under a
	training-time safety constraint, and the second stage
	--- ExpertNet --- performs fine-grained damage
	classification exclusively on samples routed as
	damaged. The safety constraint is explicit:
	$\text{Rec-Damaged} \geq 0.95$, enforced at the level
	of checkpoint selection across training epochs
	(Section~\ref{sec:sentinel}). Checkpoint selection
	maximizes precision on damaged samples subject to this
	bound --- a constrained optimization surrogate for the
	operational problem of minimizing unnecessary expert
	consultations while maintaining a hard floor on missed
	damage cases \citep{elkan2001, cannon2002}. This is not
	merely a post-hoc threshold adjustment; the constraint
	is enforced during model selection itself.
	
	The routing threshold $\tau$ is a deployment-time
	parameter. It navigates a Pareto frontier between
	expert load $\rho$ --- the fraction of samples routed
	to ExpertNet --- and Rec-Damaged. Operators can move
	along this frontier without retraining: at $\tau =
	0.5$, the system achieves $\text{Rec-Damaged} = 0.974$
	with a 13.2\% reduction in expert inference volume; at
	$\tau = 0.6$, $\text{Rec-Damaged} = 0.929$ with a
	22.6\% reduction (see Section~\ref{sec:res_sweep}). The
	selected checkpoints satisfy the safety constraint
	independently of the operating point $\tau.$
	
	The more important property is structural. ExpertNet
	always receives images drawn from a clean preprocessing
	pipeline, regardless of the corruption applied to
	Sentinel inputs. Under field and sensor corruption
	conditions, ExpertNet's diagnostic metrics ---
	F1-Expert, F1-DGT, F1-WED --- remain stable within
	measurement precision, because the applied corruption
	does not propagate to its inputs
	(Section~\ref{sec:res_robust}). What changes is $\rho$:
	the Sentinel, uncertain under degraded inputs, routes
	more samples to the Expert. The system fails toward
	consultation rather than toward confident errors. The
	flat baseline has no such mechanism; corruption
	propagates through all tasks simultaneously and
	performance degrades across the board
	\citep{hendrycks2019}.
	
	This is the paper's central claim, stated precisely:
	we identify a mismatch between objective-level risk
	encoding and system-level behavior under distribution
	shift, and show that enforcing decision structure at
	the architectural level provides a practical mechanism
	to mitigate this mismatch, yielding error-containment
	and decision-structure properties not reliably achieved
	through loss function design alone under distribution
	shift
	\citep{ovadia2019, taori2020}. More broadly, this
	suggests that under distribution shift, decision
	structure may need to be enforced at the system level
	rather than inferred from objective functions alone.
	
	The contributions are as follows.

	\begin{description}

	\item[\textbf{Architectural.}]
	A safety-aware cascaded inference framework
	that separates triage from diagnosis under
	asymmetric error costs, restructuring
	system failure modes by localizing missed
	detections at the triage stage rather than
	eliminating them globally.

	\item[\textbf{Operational.}]
	A threshold-based routing mechanism
	enabling deployment-time control of the
	safety--efficiency trade-off through
	calibrated $\tau$ selection without
	retraining. The trade-off structure between
	Rec-Healthy and Rec-Damaged is monotonic
	and predictable across evaluated settings,
	making recalibration tractable.

	\item[\textbf{Analytical.}]
	A decomposition of system behavior
	(Experiments~1--5) showing that
	architectural isolation yields error
	containment rather than intrinsic model
	robustness: ExpertNet's diagnostic
	stability under input corruption arises
	from pipeline design, not learned
	invariance, and the representational gap
	between cascade and flat baseline reduces
	to $+0.008$ F1-macro under fair evaluation.

	\item[\textbf{Empirical.}]
	Evidence on the Eyes on the Ground
	agricultural dataset demonstrating reduced
	missed damage cases (up to 54\% at
	$\tau = 0.5$ relative to a flat baseline)
	and stable expert performance under
	Sentinel-input corruption, with explicitly
	characterised trade-offs between safety,
	efficiency, and input quality.

	\end{description}
	
	The paper is organized as follows.
	Section~\ref{sec:rw} reviews related work on crop
	health assessment, multi-task learning, constrained
	classification, and cascade architectures.
	Section~\ref{sec:data} describes the Eyes on the
	Ground dataset and the label consolidation decisions
	made for this work. Section~\ref{sec:method} presents
	the CascadeCropNet architecture and training protocol.
	Section~\ref{sec:eval} defines the evaluation
	protocol. Section~\ref{sec:results} reports results,
	including the threshold sweep and robustness stress
	test. Sections~\ref{sec:discussion}
	and~\ref{sec:limitations} discuss implications and
	limitations. Section~\ref{sec:conclusion} concludes.

	
	\section{Related Work}
	\label{sec:rw}
	
	Automated crop health assessment from images has a clear
	origin story, and it is worth being honest about what
	that story actually shows. \cite{mohanty2016}
	demonstrated that a deep convolutional network trained
	on the PlantVillage dataset \citep{hughes2015} ---
	54,306 images of diseased and healthy leaves,
	photographed against controlled backgrounds --- could
	classify 26 diseases across 14 crop species with
	accuracy exceeding 99\%. The result was widely cited.
	It was also, in retrospect, a ceiling that real-world
	deployment had no realistic chance of reaching. The
	PlantVillage images are clean, centered, uniformly lit,
	and collected under conditions that bear essentially no
	resemblance to what a smallholder farmer's smartphone
	produces in the field. \cite{ramcharan2017} confronted
	this directly in their work on cassava disease detection
	in Tanzania: accuracy dropped substantially when the
	same modeling approach was applied to field-collected
	imagery, and the gap between controlled-environment
	performance and deployment performance became one of
	the defining methodological problems of the subfield.
	Subsequent work --- including \cite{barbedo2018} on
	the limits of deep learning for plant disease detection,
	\cite{atila2021} on transfer learning strategies for
	plant pathology, and \cite{picon2019} on mobile
	capture device-based crop disease classification in
	field conditions --- has probed this gap with varying
	success, but has not closed it under realistic
	deployment conditions \citep{toda2019, arsenovic2019}.
	This suggests that the limitation is not only one of
	model capacity, but of mismatch between training
	conditions and deployment environments
	\citep{kamilaris2018}.
	
	The Eyes on the Ground dataset \citep{waithaka2022} is
	positioned squarely in the hard regime. Images were
	captured by smallholder farmers across eight Kenyan
	counties using standard smartphone cameras, following a
	picture-based insurance protocol described by
	\cite{ceballos2019}. Lighting, zoom, angle, and device
	quality vary substantially across the collection.
	Labels were assigned by agronomists or trained
	professionals, with automatic labels generated by an
	in-house algorithm where manual annotation was
	unavailable. This combination of real-world image noise
	and expert-validated labels is what makes the dataset
	appropriate for safety-critical training --- you need
	both properties, and most agricultural imaging datasets
	reliably provide at most one \citep{coulibaly2019,
		sibiya2019, brahimi2017}.
	
	Multi-task learning is the natural framework when a
	single image carries multiple semantically related
	labels. The foundational argument, due to
	\cite{caruana1997}, is that shared representations
	across related tasks improve generalization by providing
	auxiliary supervision signal. More recent work has
	extended this to uncertainty-weighted loss formulations
	\citep{kendall2018} and attention-based task
	coordination \citep{liu2019}, with \cite{ruder2017}
	providing a comprehensive overview of the landscape.
	The question of which tasks benefit from joint learning
	--- and which do not --- has been studied empirically
	by \cite{standley2020}, who show that task groupings
	matter as much as architecture. In agricultural
	imaging, multi-task approaches have been used to
	jointly predict disease presence and severity
	\citep{ferentinos2018}, growth stage and damage type
	 \citep{ferentinos2018}, and various combinations of
	phenological and disturbance labels. The standard
	architecture --- shared backbone, task-specific heads,
	summed loss --- treats all tasks symmetrically. This is
	the right design when tasks are equally important and
	their labels are independent. It is the wrong design
	when one task label is only semantically meaningful
	conditional on another. Damage type --- Drought
	vs.\ Weeds --- is only a relevant prediction given that
	the crop is damaged. Training a damage-type head on
	healthy samples introduces noise: the head receives
	gradient signal from cases where its output is
	undefined by the label structure. While sufficiently
	expressive models may learn to approximate this
	conditional structure implicitly, doing so requires
	capacity to be spent modeling label dependencies that
	can instead be enforced directly \citep{standley2020}.
	Alternative approaches, such as learned gating or
	conditional computation within a single model, can
	approximate this structure, but do not guarantee that
	irrelevant samples are excluded from expert processing
	at inference time. Masked supervision, where the expert
	head is conditioned on a damage indicator and receives
	gradient only from damaged samples, is the principled
	fix. \cite{zhang2020} and \cite{guo2020} have shown
	in related hierarchical label settings that conditional
	supervision consistently outperforms symmetric
	multi-task training when label dependencies are strong.
	The cascade proposed here extends this logic from
	training to inference by enforcing conditional
	structure at the level of input distribution rather
	than gradient flow: ExpertNet never processes samples
	the Sentinel classified as Healthy, eliminating the
	problem architecturally rather than managing it through
	loss weighting.
	
	The literature on cost-sensitive learning and
	constrained classification is older and more rigorous
	than its uptake in applied agricultural ML would
	suggest. \cite{elkan2001} showed that cost-sensitive
	classification can be reduced to cost-insensitive
	classification with reweighted examples
	\citep{zadrozny2003} --- a result that is clean
	theoretically but does not address what happens when
	the input distribution shifts at deployment. The
	problem of learning from imbalanced data, closely
	related to cost-sensitive learning, has been surveyed
	by \cite{he2009} and addressed through techniques
	including synthetic oversampling \citep{chawla2002},
	threshold calibration \citep{menardi2014}, and
	re-weighting \citep{wallace2011}. Neyman-Pearson
	classification \citep{cannon2002, scott2005} frames the
	problem as constrained risk minimization: maximize
	recall on one class subject to a hard bound on the
	error rate of the other. This closely matches the
	objective structure used in the Sentinel's training,
	implemented through constraint-aware model selection
	rather than direct optimization: maximize precision on
	damaged samples subject to
	$\text{Rec-Damaged} \geq 0.95$. Selective
	classification \citep{geifman2017} offers a related
	but structurally different response: when confidence is
	insufficient, abstain and defer to a human. The
	distinction matters. Abstention removes decisions from
	the system entirely, reducing coverage
	\citep{chow1957}. Cascade routing preserves nominal
	coverage while redistributing error modes rather than
	eliminating them --- uncertain cases are not rejected
	but escalated to a specialist within the system
	boundary. These are different system philosophies, and
	conflating them obscures what the cascade actually
	does.

	In picture-based insurance schemes, the cost asymmetry
	is structural rather than parametric. A false negative
	results in claim denial for a farmer who sustained real
	losses --- a harm that cannot be reversed after the
	assessment window closes. A false positive results in
	unnecessary expert review --- an operational cost that
	scales with volume but does not cause irreversible harm
	to any individual farmer. The ratio of these costs
	depends on the specific insurance product and regional
	context \citep{barnett2007, jensen2016, ceballos2019},
	but their ordering is invariant: in any insurance scheme
	where payouts are contingent on assessment, denial of a
	legitimate claim is categorically worse than triggering
	unnecessary review.

	Cascaded classifiers have a long history in computer
	vision, beginning with the Viola-Jones face detector
	\citep{viola2001}, where a sequence of increasingly
	complex classifiers filtered easy negatives early,
	reducing total computation dramatically. Convolutional
	cascade approaches \citep{li2015} extended this to
	deep feature representations. The deep learning era
	has revisited the cascade idea primarily through the
	lens of efficiency: adaptive computation networks
	\citep{graves2016}, early-exit architectures
	\citep{bolukbasi2017}, multi-exit networks
	\citep{teerapittayanon2016}, dynamic routing
	\citep{wang2018skipnet}, spatially adaptive computation
	\citep{figurnov2017}, and multi-scale dense networks
	\citep{huang2017} all exploit the intuition that easy
	examples should not consume the same compute budget as
	hard ones \citep{panda2016}. What these systems
	optimize is throughput. What the Sentinel optimizes is
	safety. The routing criterion is not confidence in any
	prediction --- it is the probability of damage
	exceeding a threshold calibrated to a recall
	constraint. This reframes the cascade from a
	computational efficiency device into a safety
	enforcement mechanism. This distinction matters because
	optimizing for efficiency implicitly assumes symmetric
	error costs, whereas safety-constrained routing
	explicitly encodes asymmetric risk \citep{elkan2001,
		cannon2002}. To our knowledge, safety-constrained
	cascades of this form --- where routing is governed by
	a training-time recall guarantee rather than a
	confidence heuristic --- have received limited
	attention in the agricultural imaging literature.
	
	Robustness under distribution shift is a
	well-documented failure mode for convolutional networks
	generally \citep{hendrycks2019, recht2019, taori2020}
	and agricultural imaging systems specifically
	\citep{ramcharan2017, picon2019}. \cite{geirhos2019}
	showed that ImageNet-trained networks are biased toward
	texture features that are particularly vulnerable to
	certain corruption types. Corruption types that occur
	naturally in field deployment --- Gaussian noise from
	low-quality sensors, motion blur from handheld capture,
	color shift from device heterogeneity --- reliably
	degrade performance, often more than reported benchmark
	gaps would suggest \citep{hendrycks2019}. The standard
	response is data augmentation \citep{shorten2019}:
	expose the model to corrupted inputs during training
	and hope the learned representation generalizes. This
	works to a degree. What it cannot reliably do is
	protect one component of a multi-task system from
	corruption affecting another, because in a flat
	architecture all tasks share the same representation
	and therefore the same vulnerability. The reliability
	of uncertainty estimates under such shift is itself
	questionable \citep{ovadia2019, minderer2021}.
	Architectural isolation --- structurally separating the
	component that processes potentially corrupted inputs
	from the component that performs fine-grained diagnosis
	--- has received comparatively little explicit
	treatment as a robustness mechanism, particularly in
	settings where robustness to input corruption is a
	primary objective rather than a secondary effect. The
	robustness results reported in Section~\ref{sec:results}
	follow directly from this separation, not from any
	augmentation strategy applied to ExpertNet.

	
	\section{Dataset}
	\label{sec:data}
	
	The Eyes on the Ground dataset \citep{waithaka2022} was
	produced through a collaboration between ACRE Africa,
	the International Food Policy Research Institute, the
	Lacuna Fund, KALRO, BlueGreen Labs, and Dvara
	E-Registry, and is hosted on Radiant MLHub under a
	CC-BY-SA-4.0 license (DOI:
	\href{https://doi.org/10.34911/rdnt.1bs2jw}
	{10.34911/rdnt.1bs2jw}). It contains nearly 28,077 georeferenced and
	timestamped crop images collected from
	smallholder maize farms across eight counties
	in Kenya during the 2020--2021 growing seasons.
	After label consolidation and exclusion of
	classes with insufficient samples or excessive
	visual overlap (nutrient deficiency, wind,
	disease, flooding, pest, and animal damage),
	23,804 images are retained for training,
	validation, and testing. Images were
	captured by farmers using standard smartphone cameras
	following the picture-based insurance protocol of
	\cite{ceballos2019}, in which farmers photograph their
	plots at regular intervals throughout the growing
	season as a condition of insurance participation
	\citep{barnett2007, jensen2016}. Geolocation was
	recorded via mobile phone GPS; exact coordinates are
	not released, and spatial references are reported only
	at the village level using GADM36 bounding boxes to
	protect contributor privacy \citep{waithaka2022}.
	
	The label taxonomy is rich. Each image carries
	annotations along three dimensions: crop damage
	category, phenological growth stage, and damage extent.
	Damage categories include DGT (drought), WED (weed),
	WND (wind), DSE (disease), FLD (flooding), PS (pest),
	ANM (animal), ND (nutrient deficit), and GC (good
	crop). Growth stages span sowing, vegetative,
	flowering, and maturity. The full taxonomy is
	documented in the dataset property table of
	\cite{waithaka2022}, which lists both manual ML labels
	(\texttt{growth\_stage}, \texttt{damage},
	\texttt{extent}) and extended probabilistic labels
	(\texttt{drought\_probability},
	\texttt{disturbance\_weed},
	\texttt{disturbance\_drought}, and others) derived from
	independent automatic classifiers. All labels were
	assigned by agronomists or trained professionals where
	manual annotation was feasible; an in-house machine
	learning algorithm supplied automatic labels where
	manual review was unavailable, with automatic label
	probabilities derived from independent models and
	therefore not constrained to sum to unity. This
	labeling pipeline is imperfect by design --- it
	reflects the operational realities of large-scale data
	collection in the field --- and the mixed
	manual-automatic provenance should be understood as a
	property of the dataset, not a defect to be corrected
	away. Because automatic labels are produced by
	independent models, the resulting noise is structured
	rather than purely random \citep{zhang2020}, which may
	affect both class boundaries and calibration
	\citep{minderer2021}. The impact of this mixed
	provenance is implicitly evaluated through the
	robustness and generalization behavior reported in
	Section~\ref{sec:results}.
	
	For this work, the full label taxonomy is consolidated
	into three binary or categorical targets aligned with
	the cascade's two-stage decision structure. Health
	status collapses to a binary label: Healthy (GC)
	versus Damaged (DGT + WED). This restriction defines a
	controlled setting for evaluating asymmetric decision
	policies under class imbalance
	\citep{he2009, menardi2014}; evaluating the cascade
	under the full multi-class taxonomy is an important
	direction for future work. Nutrient deficiency cases
	are excluded from the damage severity task --- the
	class contains too few samples for reliable supervised
	learning and preliminary inspection indicates
	substantial visual overlap with drought stress, making
	inclusion more likely to introduce label ambiguity than
	useful signal \citep{saleem2019}. Growth stage is
	consolidated from four phenological classes to two:
	Early (Sowing + Vegetative) and Late (Flowering +
	Maturity). This aggregation may discard finer-grained
	temporal cues, but provides a more stable supervisory
	signal given class frequencies, while preserving the
	agronomically meaningful distinction between vegetative
	and reproductive phases. 
    
    The cascade uses a hierarchical label schema with three targets
    --- \texttt{bin\_label}, \texttt{expert\_label},
    \texttt{growth\_label} --- while the flat baseline uses a
    3-class damage label (Healthy, Drought, Weeds) alongside
    \texttt{growth\_label}. These two schemas are applied to the
    same underlying image collection but encode different
    assumptions about task structure, and comparison between
    systems trained under each schema requires care.
	
\begin{figure*}[htbp]
		\centering
		\includegraphics[width=\textwidth]{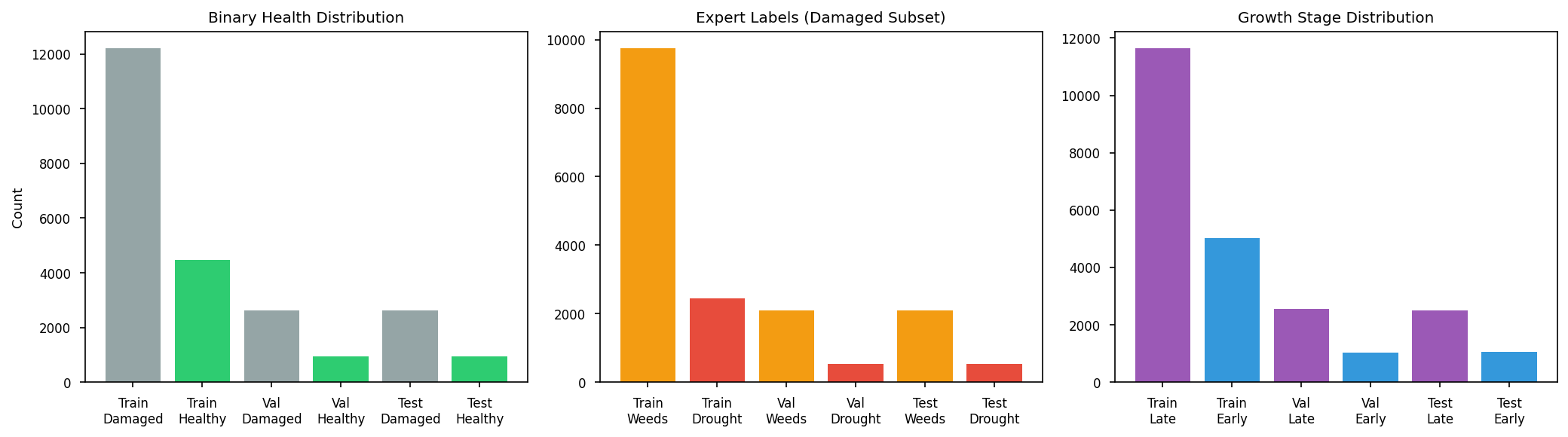}
		\caption{Class distribution across train, validation,
			and test splits. \textbf{Left:} Binary health
			distribution (Damaged vs.\ Healthy). The 73.3\%
			Damaged rate is consistent across all three splits,
			reflecting stratified sampling on the damage label.
			\textbf{Center:} Expert label distribution within
			the damaged subset. Weeds (WED) account for 79.9\%
			of damaged samples versus 20.1\% Drought (DGT),
			producing a 4:1 class imbalance addressed via
			FocalLoss \citep{lin2017} in ExpertNet training.
			\textbf{Right:} Growth stage distribution (Late
			vs.\ Early). Late growth (Flowering + Maturity)
			accounts for 69.9\% of all images. Data from
			\cite{waithaka2022}.}
		\label{fig:eda}
	\end{figure*}

	The dataset is partitioned into train, validation, and
	test splits at a 70/15/15 ratio using fixed CSV files,
	with splits performed at the image level and verified
	to have no overlap across sets (Table~\ref{tab:splits}).
	Because plot-level identifiers are not available,
	independence at the field level cannot be confirmed;
	this may introduce residual correlation across splits
	due to repeated observations of the same plots over
	time \citep{ceballos2019}, and reported performance
	metrics should be interpreted with this limitation in
	mind. Splits are stratified on the damage label to
	preserve class frequencies. The resulting distribution is summarized in
	Table~\ref{tab:splits}. The training set contains
	16,662 images; validation and test sets contain 3,571
	each. Three levels of class imbalance are present and
	require explicit treatment \citep{he2009, chawla2002},
	all of which are visible in the class distribution
	plots of Figure~\ref{fig:eda}. At the health level,
	73.3\% of images are Damaged and 26.7\% are Healthy
	--- a roughly 2.7:1 ratio that persists consistently
	across all three splits (Figure~\ref{fig:eda}, left
	panel). Among damaged images, Weeds account for 79.9\%
	of cases and Drought for 20.1\%, a 4:1 imbalance that
	represents the primary challenge for the Expert model
	(Figure~\ref{fig:eda}, center panel). At the growth
	stage level, Late growth (Flowering + Maturity)
	accounts for 69.9\% of images versus 30.1\% Early ---
	a milder imbalance that requires no special weighting
	beyond standard cross-entropy
	(Figure~\ref{fig:eda}, right panel).
	
	\begin{table*}[htbp]
		\centering
	\caption{Dataset split statistics. Health, damage
		type, and growth stage distributions are reported as
		percentages within each split. Source:
		\cite{waithaka2022}; split proportions computed
		from the stratified 70/15/15 partition described in
		Section~\ref{sec:data}.}
		\label{tab:splits}
		\small
		\begin{tabular}{lrrrrrrr}
			\toprule
			\textbf{Split} & \textbf{Total} &
			\textbf{Healthy} & \textbf{Damaged} &
			\textbf{Drought} & \textbf{Weeds} &
			\textbf{Early} & \textbf{Late} \\
			\midrule
			Train & 16{,}662 & 26.7\% & 73.3\% &
			14.8\% & 58.5\% & 30.1\% & 69.9\% \\
			Val   & 3{,}571  & 26.8\% & 73.2\% &
			14.7\% & 58.5\% & 28.8\% & 71.2\% \\
			Test  & 3{,}571  & 26.7\% & 73.3\% &
			14.8\% & 58.5\% & 29.7\% & 70.3\% \\
			\midrule
			\textbf{Total} & \textbf{23{,}804} & 26.7\% & 73.3\%
			& 14.8\% & 58.5\% & 29.9\% & 70.1\% \\
			\bottomrule
		\end{tabular}
	\end{table*}
	
	\begin{table*}[htbp]
		\centering
		\caption{SentinelNet (V13) and flat baseline
			(V11) training configuration.}
		\label{tab:sentinel_config}
		\small
		\begin{tabular}{lll}
			\toprule
			\textbf{Setting} & \textbf{V13 Sentinel}
			& \textbf{V11 Baseline} \\
			\midrule
			Epochs          & 30       & 20 \\
			Batch size      & 32       & 16 \\
			Learning rate   & 1e-4     & 1e-4 \\
			Weight decay    & 0.01     & 0.01 \\
			LR schedule     & ReduceLROnPlateau
			& ReduceLROnPlateau \\
			Patience/factor & 3 / 0.5  & 3 / 0.5 \\
			Backbone warmup & 3 epochs & 5 epochs \\
			Class weights   & [1.37, 3.70] & None \\
			Selection criterion & max(Safety Score) &
			max(0.6·F1-D + 0.4·F1-G) \\
			\bottomrule
		\end{tabular}
	\end{table*}
	
	\begin{table*}[htbp]
		\centering
		\caption{ExpertNet (V14) training
			configuration.}
		\label{tab:expert_config}
		\small
		\begin{tabular}{ll}
			\toprule
			\textbf{Setting} & \textbf{V14 Expert} \\
			\midrule
			Epochs           & 25 \\
			Batch size       & 16 \\
			Learning rate    & 1e-4 \\
			Weight decay     & 0.05 \\
			LR schedule      & CosineAnnealingLR \\
			Backbone warmup  & 5 epochs \\
			Progressive sampler & Yes \\
			Corruption augmentation &
			ControlledCorruption + RandomNoise \\
			& (50\% clean, 20\% blur, 30\%
			colour jitter, \\
			& $\sigma=0.08$, $p=10\%$) \\
			Selection criterion &
			max(0.7·F1-Exp + 0.3·Rec-H) \\
			\bottomrule
		\end{tabular}
	\end{table*}
	
	These three imbalances are not independent problems to
	be managed separately. They interact. The Sentinel
	faces a dual challenge: identifying the minority
	Healthy class while maintaining a hard recall
	constraint on the majority Damaged class
	\citep{cannon2002, scott2005}. The Expert must
	distinguish Drought from Weeds in a 4:1 imbalanced
	regime, on a subset of images that has already been
	filtered by the Sentinel's routing decision. Imbalance
	handling is therefore specific to each stage: the
	Sentinel uses inverse-frequency class weights of
	$[1.37,\ 3.70]$ for Damaged and Healthy respectively,
	combined with ClassBalancedFocalLoss \citep{cui2019};
	the Expert uses FocalLoss with $\gamma = 2.0$
	\citep{lin2017} on the damage-type head, combined with
	a progressive sampler that upweights damaged samples
	during training. These choices follow standard practice
	for imbalanced classification \citep{he2009,
		wallace2011} and are applied consistently across models
	to reduce confounding from class imbalance, although
	differences in optimization dynamics across
	architectures may still persist \citep{kendall2018}.
	Growth stage prediction at both stages uses standard
	binary cross-entropy without reweighting, given the
	milder imbalance. The full training configuration for
	all three models is described in
	Section~\ref{sec:method}.
	
	One property of this dataset deserves emphasis beyond
	the class statistics. The images are genuinely
	difficult. Smartphone cameras vary in sensor quality,
	lens characteristics, and automatic exposure behavior
	--- a consequence of the mobile phone GPS collection
	protocol documented in \cite{waithaka2022}. Farmers
	photograph their plots from varying distances and
	angles. Lighting conditions range from direct midday
	sun --- which washes out color information --- to
	overcast conditions that flatten contrast. Motion blur
	from handheld capture is common \citep{shorten2019}.
	These are not pathological edge cases; they are the
	baseline distribution \citep{hendrycks2019}, and
	therefore constitute a natural domain shift relative to
	controlled imaging conditions \citep{recht2019,
		taori2020}. The stress test reported in
	Section~\ref{sec:results} evaluates robustness under
	this shift directly, using the three corruption
	conditions visualized in Figure~\ref{fig:stress}.
	
	
	\section{Method}
	\label{sec:method}
	
	The design of CascadeCropNet is motivated by a single
	observation: the decision structure of crop damage
	assessment is hierarchical by nature, and a flat
	classifier that ignores this structure pays a cost in
	both safety and robustness \citep{standley2020,
		hendrycks2019}. Damage type is only a meaningful
	prediction conditional on damage presence. Growth stage
	is an auxiliary signal relevant throughout. These are
	not implementation details --- they are properties of
	the label space that should be reflected in the
	architecture, not managed through loss weighting after
	the fact \citep{caruana1997, ruder2017}.
	
	\subsection{Design Rationale}
	\label{sec:rationale}
	
	We cast crop damage triage as a constrained decision
	problem, implemented via constrained model selection
	over training checkpoints. Let $\mathbf{x}$ denote an
	input image and $y \in \{\text{Healthy, Damaged}\}$
	its binary health label. The Sentinel learns a scoring
	function $f(\mathbf{x}) = P(\text{Damaged}|\mathbf{x})$,
	and a threshold $\tau$ is selected on the validation
	set such that the resulting classifier maximizes
	$\text{Safety Score}$ subject to
	$\text{Rec-Damaged} \geq 0.95$. Formally:
	
	
	\begin{equation}
	\tau^{*} = \arg\max_{\tau}\;
	\text{Safety Score}(\tau)
	\quad \text{s.t.} \quad
	\text{Rec-Damaged}(\tau) \geq 0.95
	\label{eq:constraint}
	\end{equation}
	
	Precision serves as a tractable proxy for operational
	efficiency --- minimizing unnecessary expert
	consultations --- while the recall constraint encodes
	the asymmetric cost floor: the cost of missing damage
	exceeds the cost of unnecessary review
	\citep{elkan2001, cannon2002}. This formulation can
	be interpreted as enforcing a minimum safety constraint
	while optimizing operational efficiency within the
	feasible region --- a form of structured risk control
	in which the safety floor is non-negotiable and
	efficiency is maximized subject to it
	\citep{scott2005, geifman2017}. This is not a full
	cost model \citep{zadrozny2003}. It is a principled
	surrogate that makes the asymmetry explicit and
	enforceable at the level of model selection rather than
	gradient optimization.
	
	Expert load $\rho$ --- the fraction of samples routed
	to ExpertNet --- is reported directly as the primary
	efficiency metric, alongside Rec-Damaged as the primary
	safety metric. The choice of precision as the secondary
	objective reflects its operational alignment: a
	Sentinel that produces fewer false positives routes
	fewer healthy samples to the Expert, reducing $\rho$
	directly. Precision and $\rho$ are related but not
	identical; reporting $\rho$ explicitly avoids
	conflating the two \citep{davis2006}. The constraint
	and objective together define a Pareto frontier, with
	x-axis expert load $\rho$ and y-axis Rec-Damaged
	defining the operating space. The threshold $\tau$
	navigates this frontier at deployment time without
	retraining, as visualized in
	Figure~\ref{fig:threshold} (left panel).
	
	\subsection{SentinelNet}
	\label{sec:sentinel}

	SentinelNet (V13) is the first stage of the cascade.
	Its primary task is binary health discrimination; its
	routing output determines which samples ExpertNet
	processes. The backbone is MobileNetV3-Large
	\citep{howard2019} pretrained on ImageNet
	\citep{deng2009}, with approximately 4.2 million
	parameters. Input resolution is $224 \times 224$
	pixels. The shared feature extractor feeds into a
	fully connected layer ($960 \to 1024$) followed by
	batch normalization and ReLU activation, from which
	three task-specific heads branch: a binary health head
	($1024 \to 256 \to 2$, Healthy/Damaged), an auxiliary
	damage-type head ($1024 \to 256 \to 2$,
	Drought/Weeds), and an auxiliary growth stage head
	($1024 \to 256 \to 2$, Early/Late). All heads use
	ReLU activation and Dropout(0.2) before the final
	linear layer.
	
	The training loss combines three terms with explicit
	weighting:
	
	\begin{equation}
		\mathcal{L}_{\text{S}} =
		5 \cdot \mathcal{L}_{\text{bin}}^{\text{CBFocal}} +
		1 \cdot \mathcal{L}_{\text{exp}}^{\text{CE}} +
		0.1 \cdot \mathcal{L}_{\text{gro}}^{\text{CE}}
	\end{equation}
	
	The binary health loss uses ClassBalancedFocalLoss
	\citep{cui2019} with inverse-frequency class weights
	$[1.37,\ 3.70]$ for Damaged and Healthy respectively,
	derived from the training set frequencies reported 
	in Table~\ref{tab:splits}. The heavy weighting on the
	binary term ($\times 5$) reflects the primacy of health
	triage in the cascade's decision structure.

The auxiliary growth stage head contribution to the
total loss is zeroed after the first five training
epochs ($\mathcal{L}_{\text{gro}} = 0$ for
$\text{epoch} \geq 5$), concentrating gradient signal
on the binary health and auxiliary damage-type tasks
as training progresses \citep{ruder2017}. The
auxiliary damage-type head receives standard
cross-entropy supervision throughout all 30 training
epochs at weighting coefficient 1.0, providing
continuous signal for the shared representation to
encode damage-relevant visual features
\citep{caruana1997}.
	
	Training runs for 30 epochs with AdamW
	\citep{loshchilov2019} ($\text{lr} = 1\times10^{-4}$,
	weight decay $= 0.01$), gradient clipping at
	$\text{max\_norm} = 1.0$, and ReduceLROnPlateau
	scheduling (patience $= 3$, factor $= 0.5$). The
	backbone is frozen for the first three epochs with
	batch normalization in evaluation mode to stabilize
	early feature extraction \citep{howard2019}. 
We define \emph{Safety Score} as the precision
on the Damaged class evaluated at the
highest-confidence threshold satisfying both
Rec-Damaged $\geq 0.95$ and Rec-Healthy $\geq
0.40$ simultaneously. Epochs where no such
threshold exists receive Safety Score $= 0$
and are ineligible for selection. Checkpoint
selection maximizes Safety Score on the
validation set at each epoch. The threshold $\tau$ is
	tuned per epoch on the validation set by sweeping
	$P(\text{Damaged}|\mathbf{x})$ over a dense grid and
	selecting the value that maximizes the Safety Score
	subject to the constraints in
	Equation~\ref{eq:constraint}. This mechanism enforces
	satisfaction of the recall constraint on the validation
	set, avoiding approximation errors introduced by
	surrogate penalty terms \citep{cannon2002, scott2005}.
	Training dynamics are shown in 
	Figure~\ref{fig:sentinel}, showing loss curves, recall
	trajectories for Rec-Healthy and Rec-Damaged, and the
	optimal $\tau$ evolution across 30 epochs.
	
	\subsection{ExpertNet}
	\label{sec:expert}
	
	ExpertNet (V14) is the second stage. It receives only
	samples routed by the Sentinel and performs
	fine-grained classification between Drought and Weeds.
	The architecture mirrors SentinelNet --- MobileNetV3-Large
	backbone \citep{howard2019}, shared fully connected
	layer, three task-specific heads --- but operates at
	higher input resolution ($256 \times 256$ pixels) to
	capture the finer visual detail that distinguishes
	drought stress from weed competition
	\citep{fuentes2017, toda2019}. The damage-type head is
	the primary output; the binary health head and growth
	head serve as auxiliary tasks. Training dynamics for
	V14 are shown in Figure~\ref{fig:expert}, showing
	loss curves, per-class diagnostic F1 scores (F1-DGT
	and F1-WED), and the composite HP-Score across 25
	epochs.
	
	The training loss weights the damage-type task heavily:
	
	\begin{equation}
		\mathcal{L}_{\text{E}} =
		1 \cdot \mathcal{L}_{\text{bin}}^{\text{CE,weighted}}
		+ 3 \cdot \mathcal{L}_{\text{exp}}^{\text{Focal}
			(\gamma=2)} +
		0.1 \cdot \mathcal{L}_{\text{gro}}^{\text{CE}}
	\end{equation}
	
	FocalLoss with $\gamma = 2.0$ \citep{lin2017} on the
	damage-type head addresses the 4:1 Weeds/Drought
	imbalance documented in Section~\ref{sec:data} by
	down-weighting easy majority-class examples during
	training \citep{cui2019}. A progressive sampler further
	upweights damaged samples in each batch \citep{he2009}.
	Corruption augmentation is applied during training ---
	ControlledCorruption (50\% clean, 20\% Gaussian blur,
	30\% colour jitter) combined with RandomNoise
	($\sigma = 0.08$, applied with probability 10\%)
	\citep{shorten2019} --- to improve the Expert's
	robustness to inputs that may arrive with residual
	preprocessing artifacts. This augmentation is applied
	to ExpertNet's training distribution, not to Sentinel
	inputs at inference.
	
	Training runs for 25 epochs with AdamW
	\citep{loshchilov2019} ($\text{lr} = 1\times10^{-4}$,
	weight decay $= 0.05$) and CosineAnnealingLR
	scheduling. The backbone is frozen for the first five
	epochs. Checkpoint selection maximizes the composite
	score $0.7 \cdot \text{F1-Expert} +
	0.3 \cdot \text{Rec-Healthy}$ on the validation set,
	with the best checkpoint identified at epoch 7
	($\text{combined\_hp} = 0.8278$). Total training time across
	all three models is approximately 20.7 hours: 4.2
	hours for V11, 10.0 hours for V13 SentinelNet, and 6.5
	hours for V14 ExpertNet, all on GPU with a fixed random
	seed of 42 for reproducibility.
	
	\subsection{Cascade Routing and Architectural
		Isolation}
	\label{sec:routing}
	
	At inference, the cascade operates as follows. Every
	input image is first processed by SentinelNet at
	$224 \times 224$ pixels, producing
	$P(\text{Damaged}|\mathbf{x})$. If
	$P(\text{Damaged}|\mathbf{x}) \geq \tau$, the image is
	routed to ExpertNet at $256 \times 256$ pixels; the
	Expert's damage-type output (Drought or Weeds) is used
	as the final prediction. If
	$P(\text{Damaged}|\mathbf{x}) < \tau$, the image is
	predicted Healthy and ExpertNet is never invoked.
	Growth stage prediction is always drawn from the
	Sentinel, regardless of routing. The fraction of
	samples routed to ExpertNet is $\rho$, the expert load.
	This inference procedure is illustrated in
	Figure~\ref{fig:threshold}, which also shows the full
	threshold sweep across
	$\tau \in \{0.3, 0.4, 0.5, 0.6, 0.7, 0.8\}$ on the
	validation set.
	
	We define \emph{architectural isolation} formally as
	follows: ExpertNet's input distribution is invariant to
	corruptions applied to Sentinel inputs, under a fixed
	preprocessing pipeline. This holds structurally.
	Corruption affects $P(\text{Damaged}|\mathbf{x})$ and
	therefore $\rho$ --- the routing composition --- but
	does not affect the pixel values passed to ExpertNet,
	which are always drawn from the clean preprocessing
	pipeline applied to the original source image.
	Architectural isolation therefore guarantees
	\emph{input quality invariance}, not
	\emph{selection stability}. These are two distinct
	properties \citep{ovadia2019}. Under corruption, the
	set of samples reaching ExpertNet changes --- the
	Sentinel routes more aggressively when uncertain --- but
	each routed sample arrives at ExpertNet with its
	representation intact. This routing-induced shift may
	also alter the conditional difficulty of the expert
	task, as samples routed under high Sentinel uncertainty
	are not drawn from the same distribution as clean
	training examples \citep{taori2020}. The robustness
	results reported in Section~\ref{sec:results} follow
	from the input quality invariance property; the
	deployment caveat about routing-induced distribution
	shift follows from the selection stability property.
	
	This distinction from abstention-based approaches is
	structural, not incidental \citep{chow1957,
		geifman2017}. Selective classifiers reduce coverage by
	removing low-confidence decisions from the system. The
	cascade preserves nominal coverage while redistributing
	error modes rather than eliminating them
	\citep{cannon2002}: every input receives a final
	prediction, but the decision pathway depends on the
	Sentinel's confidence. Uncertain samples are escalated
	within the system rather than deferred outside it.
	
	\subsection{Masked Supervision}
	\label{sec:masking}
	
	The damage-type head in both SentinelNet and ExpertNet
	is supervised exclusively on damaged samples. Formally,
	the expert loss term is conditioned on the damage
	indicator:
	
	\begin{equation}
		\mathcal{L}_{\text{exp}} =
		\mathbf{1}_{[y \in \{\text{Drought, Weeds}\}]}
		\cdot \ell\bigl(f_{\text{exp}}(\mathbf{x}),\,
		y_{\text{exp}}\bigr)
	\end{equation}
	
	where $\ell$ is the appropriate loss function and
	$f_{\text{exp}}$ is the damage-type head output.
	Without this masking, the expert head receives gradient
	signal from healthy samples --- cases where the
	damage-type label is structurally undefined ---
	introducing inconsistent supervision signals for
	samples where the damage-type label carries no meaning
	\citep{standley2020, zhang2020}. Masked supervision
	enforces the conditional structure of the label space
	at the gradient level; the cascade enforces it at the
	input distribution level at inference. These two
	mechanisms are complementary and operate at different
	stages of the learning and inference pipeline
	\citep{ruder2017, liu2019}.
	
	\subsection{Flat Baseline}
	\label{sec:baseline}
	
\begin{figure*}[htbp]
		\centering
		\includegraphics[width=\textwidth]{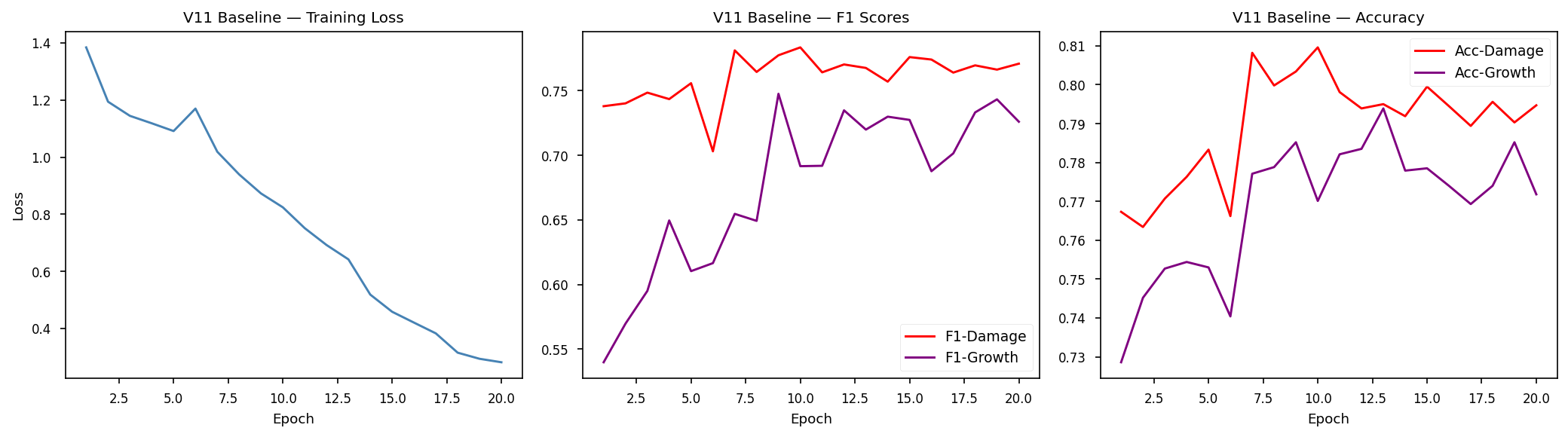}
		\caption{V11 flat baseline training dynamics across
			20 epochs.
			\textbf{Left:} Training loss decreasing smoothly,
			converging by epoch 9.
			\textbf{Center:} F1-Damage (3-class macro) and
			F1-Growth (binary macro) trajectories. F1-Damage
			plateaus around 0.77--0.78 from epoch 9 onward,
			consistent with capacity limits of a flat
			architecture attempting to model conditional label
			dependencies without structural enforcement
			\citep{standley2020}.
			\textbf{Right:} Accuracy on damage and growth tasks.
			Best checkpoint: epoch 9, combined score $= 0.7654$.}
		\label{fig:v11}
	\end{figure*}

	The flat baseline (V11, BaselineNetV11) uses the same
	MobileNetV3-Large backbone \citep{howard2019} with two
	task-specific heads: a 3-class damage head (Healthy,
	Drought, Weeds) and a binary growth head (Early, Late).
	Input resolution is $224 \times 224$ pixels. The
	training loss is an equal-weighted sum of cross-entropy
	losses on both tasks. No routing, no safety constraint,
	no masked supervision. The selection criterion
	maximizes $0.6 \cdot \text{F1-Damage} +
	0.4 \cdot \text{F1-Growth}$ on the validation set.
	Training runs for 20 epochs with AdamW
	\citep{loshchilov2019}
	($\text{lr} = 1\times10^{-4}$, weight decay $= 0.01$)
	and ReduceLROnPlateau scheduling, with a five-epoch
	backbone warmup, converging at epoch 9 as shown
	in Figure~\ref{fig:v11}.
	The baseline is deliberately minimal: its purpose is
	to isolate the effect of the cascade architecture and
	safety constraint against a comparable backbone
	\citep{howard2019}. The absence of stronger flat
	baselines should be interpreted as a limitation of the
	current study rather than evidence of superiority over
	all flat approaches.
	
	
	\begin{figure*}[htbp]
		\centering
		\includegraphics[width=\textwidth]{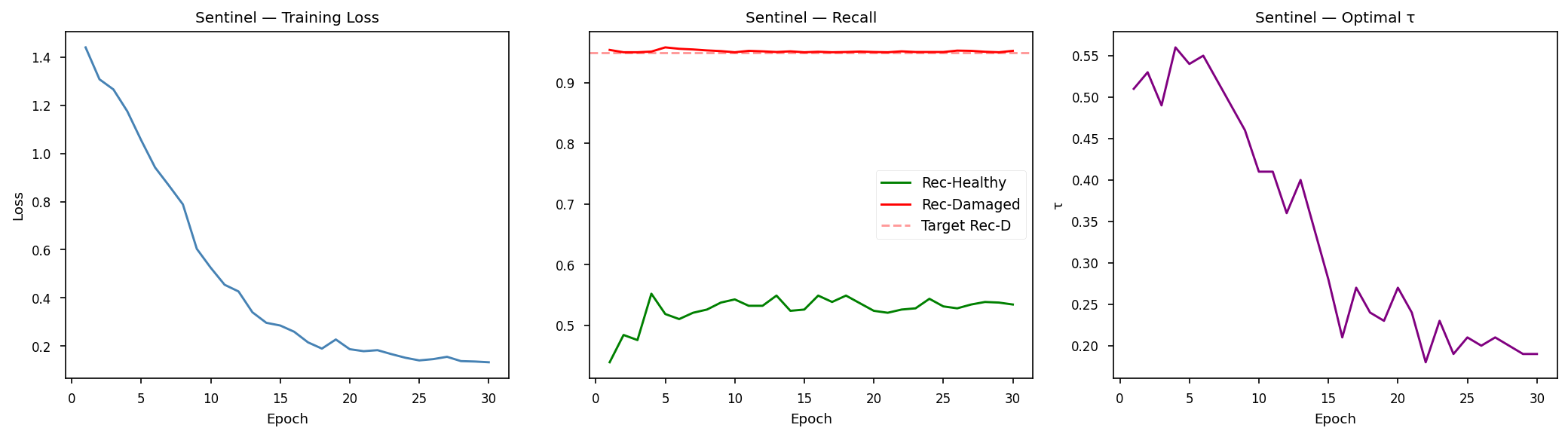}
		\caption{SentinelNet (V13) training dynamics across
			30 epochs.
			\textbf{Left:} Training loss curve showing smooth
			convergence.
			\textbf{Center:} Recall trajectories for
			Rec-Damaged (red, maintained $\geq 0.95$ throughout)
			and Rec-Healthy (green), with the target Rec-Damaged
			$= 0.95$ constraint shown as a dashed line. The
			safety constraint is satisfied at every epoch under
			appropriate threshold selection.
			\textbf{Right:} Optimal $\tau$ evolution across
			training epochs, drifting from $\tau = 0.56$ at
			epoch 4 to $\tau = 0.19$ at epoch 30, reflecting
			the model's increasing confidence in damage
			predictions. Best checkpoint: epoch 4, Safety Score
			$= 0.8532$.}
		\label{fig:sentinel}
	\end{figure*}
	
	
	\begin{figure*}[htbp]
		\centering
		\includegraphics[width=\textwidth]{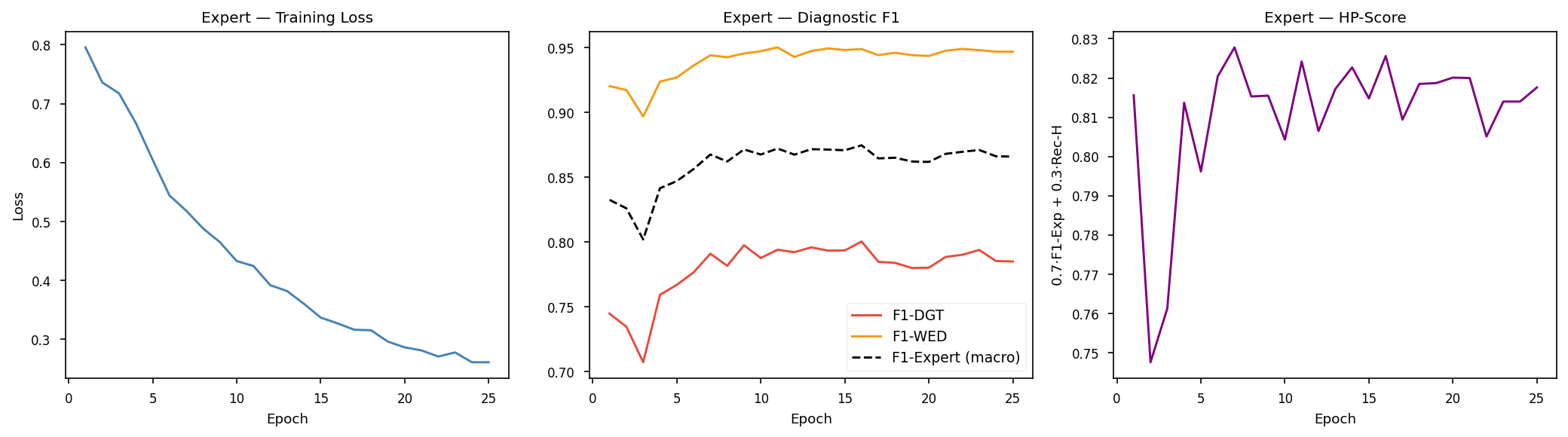}
		\caption{ExpertNet (V14) training dynamics across
			25 epochs.
			\textbf{Left:} Training loss converging smoothly
			throughout.
			\textbf{Center:} Per-class diagnostic F1 scores.
			F1-WED (Weeds, orange) rises steeply and stabilizes
			near 0.95 by epoch 5. F1-DGT (Drought, red) climbs
			more slowly and plateaus near 0.79, consistent with
			the 4:1 Weeds/Drought imbalance \citep{lin2017,
				cui2019} and feature-space overlap between the two
			damage classes \citep{saleem2019}. F1-Expert macro
			(dashed black) stabilizes around 0.87.
			\textbf{Right:} Composite HP-Score
			($0.7 \cdot \text{F1-Expert} +
			0.3 \cdot \text{Rec-Healthy}$) across epochs.
			Best checkpoint: epoch 7,
			$\text{combined\_hp} = 0.8278$.}
		\label{fig:expert}
	\end{figure*}
	
	\subsection*{Reproducibility}
	\label{sec:repro}
	
	All experiments use a fixed random seed of 42
	applied to Python \texttt{random}, NumPy,
	PyTorch, and \texttt{cudnn.deterministic=True}.
	The framework is PyTorch with torchvision,
	using MobileNetV3-Large pretrained on ImageNet
	\citep{deng2009, howard2019}. Dataset splits
	are fixed CSV files with no overlap between
	train, validation, and test sets. Thresholds
	$\tau \in \{0.3, 0.4, 0.5, 0.6, 0.7, 0.8\}$
	are evaluated post-hoc on the validation set
	after training is complete; threshold selection
	never touches the test set. Training flags
	default to checkpoint loading
	(\texttt{TRAIN\_FLAT}, \texttt{TRAIN\_SENTINEL},
	\texttt{TRAIN\_EXPERT} all default
	\texttt{False}). Total training time across all
	three models is approximately 20.7 hours: 4.2h
	for V11, 10.0h for V13 SentinelNet, and 6.5h
	for V14 ExpertNet, all on GPU. Model
	checkpoints are saved with
	\texttt{model\_state\_dict}, threshold,
	Safety Score, and epoch metadata. The complete
	training configuration is documented in
	Tables~\ref{tab:sentinel_config}
	and~\ref{tab:expert_config}. Code and
	checkpoints are available from the
	corresponding author upon reasonable request.
	
	
	\section{Evaluation Protocol}
	\label{sec:eval}
	
	Evaluation in a two-stage cascade requires choices that
	a flat classifier does not face. The most consequential
	is this: should ExpertNet be evaluated on samples that
	the Sentinel actually routes to it, or on the full set
	of truly damaged samples? These are not the same
	population. In deployment, only sentinel-routed samples
	reach ExpertNet --- the Expert never sees images the
	Sentinel classified as Healthy, regardless of their
	true label. Evaluating on routed samples would confound
	two distinct sources of error: Sentinel routing
	mistakes and ExpertNet diagnostic mistakes. Separating
	them requires a deliberate protocol choice
	\citep{davis2006, powers2011}.
	
	Throughout this work, ExpertNet is evaluated on truly
	damaged samples --- the full set of images carrying a
	Drought or Weeds label in the ground truth, irrespective
	of how the Sentinel routed them. This isolates
	ExpertNet's intrinsic diagnostic capacity from triage
	performance, enabling controlled comparison of what the
	Expert actually learned. For
	completeness, evaluation under actual Sentinel routing
	is also captured in the end-to-end cascade metrics ---
	Rec-Damaged, Rec-Healthy, F1-bin, and $\rho$ ---
	ensuring that both intrinsic and deployed Expert
	performance are observable. End-to-end cascade
	performance constitutes the primary evaluation, while
	decomposed Expert metrics serve diagnostic
	interpretation. The trade-off is transparency about
	what each level measures: F1-Expert, F1-DGT, and
	F1-WED reflect the Expert's representational quality;
	end-to-end metrics reflect system behavior under actual
	routing. In deployment, if routing is systematically
	biased --- over-routing certain crop varieties,
	under-routing early-stage damage --- deployed Expert
	performance will differ from the decomposed figures
	reported here \citep{ovadia2019}. This separation
	assumes that representation quality generalizes across
	routing-induced distribution shifts \citep{taori2020}.
	That assumption remains empirically unverified and
	constitutes an open problem for follow-on work.
	
	The threshold sweep is conducted on the validation set
	($n = 3{,}571$) across six operating points:
	$\tau \in \{0.3, 0.4, 0.5, 0.6, 0.7, 0.8\}$,
	as described in Section~\ref{sec:rationale}. At each
	$\tau$, expert load $\rho$ and all cascade metrics are
	computed. The Pareto frontier between $\rho$ (x-axis)
	and Rec-Damaged (y-axis) is visualized in
	Figure~\ref{fig:threshold} (left panel), alongside the
	full metric sweep across $\tau$ values in
	Figure~\ref{fig:threshold} (right panel). While both
	the cascade and the flat baseline are subject to
	validation-dependent model selection, the cascade
	introduces an additional operating parameter $\tau$,
	which may increase sensitivity to selection effects
	relative to V11 \citep{menardi2014}. This asymmetry is
	acknowledged; reported metrics should be interpreted
	accordingly.
	
	The robustness stress test is conducted on the held-out
	test set ($n = 3{,}571$) under three input corruption
	conditions applied exclusively to Sentinel inputs.
	ExpertNet always receives images from the clean
	preprocessing pipeline, consistent with the
	architectural isolation property defined in
	Section~\ref{sec:routing}. The three conditions are:
	\textbf{Clean} (standard resize and normalize),
	\textbf{Field} (ColorJitter with brightness and
	contrast factor 0.4, combined with GaussianBlur with
	$\sigma \in [0.1, 2.0]$), and \textbf{Sensor}
	(additive Gaussian noise with $\sigma = 0.08$ applied
	post-normalization), as described in the Reproducibility subsection above. These conditions simulate
	the dominant corruption types present in the Eyes on
	the Ground dataset \citep{waithaka2022}: device
	heterogeneity and exposure variation (Field), and
	sensor noise from low-quality smartphone cameras
	(Sensor) \citep{hendrycks2019, shorten2019}. Corruption
	is applied to Sentinel inputs only --- not to ExpertNet
	inputs --- because this reflects the intended
	deployment design: in a real pipeline, the Expert model
	operates on images that have passed through a
	controlled server-side preprocessing stage, while the
	Sentinel operates closer to the raw capture device
	\citep{ceballos2019}. The flat baseline, which has no
	such architectural separation, is evaluated under
	identical corruption conditions applied uniformly to
	its single model. This asymmetry is not an evaluation
	artifact --- it is a direct consequence of the
	architectural difference being tested.
	
	The cascade decomposes a single harder joint
	classification problem into two simpler conditional
	ones \citep{standley2020}. A reviewer might reasonably
	ask whether performance gains simply reflect an easier
	subtask rather than a genuine system improvement. The
	response is structural: while each subtask is
	individually simpler, the overall system is evaluated
	on the full input distribution --- every image receives
	a final prediction, and end-to-end metrics capture the
	complete decision chain including routing errors. Task
	decomposition is the architectural contribution; the
	end-to-end evaluation is what makes the gains
	meaningful \citep{elkan2001, cannon2002}.
	
	A note on split usage. The threshold sweep uses the
	validation set; the stress test uses the test set.
	$\tau$ is a validation-time hyperparameter ---
	selecting it on the test set would constitute
	evaluation leakage \citep{menardi2014} --- while the
	stress test is a post-hoc robustness evaluation applied
	to a fixed, already-selected model. The practical
	consequence is that threshold sweep metrics and stress
	test metrics are not directly comparable as absolute
	numbers across splits. Both sets of results are
	reported with their respective split origins made
	explicit. Harmonizing these evaluations to a single
	held-out split is a methodological improvement left for
	future work, as noted in Section~\ref{sec:limitations}.
	
	The following metrics are reported throughout
	this work. \textbf{Rec-Damaged}:
	recall on the Damaged class, the primary safety metric
	and the quantity constrained during Sentinel training
	\citep{cannon2002, scott2005}. \textbf{Rec-Healthy}:
	recall on the Healthy class, the explicit operational
	cost of the safety floor \citep{elkan2001}.
	\textbf{F1-bin}: macro F1 on the binary health task at
	the cascade level, capturing end-to-end routing
	performance \citep{powers2011}. \textbf{F1-Expert}:
	macro F1 on the damage-type task (Drought vs.\ Weeds),
	computed on truly damaged samples. \textbf{F1-DGT} and
	\textbf{F1-WED}: per-class F1 scores for Drought and
	Weeds respectively, drawn from the Expert's confusion
	matrix as shown in Figure~\ref{fig:confusion}. \textbf{F1-Growth}: macro
	F1 on the binary growth stage task, always predicted
	by the Sentinel \citep{powers2011}. $\rho$:
	expert load, the fraction of samples routed to
	ExpertNet, reported as the primary efficiency metric.
	All F1 scores are macro-averaged \citep{davis2006,
		powers2011}. Confusion matrices for all three tasks at
	$\tau = 0.5$ are provided in
	Figure~\ref{fig:confusion}. Training dynamics for all
	three models are documented in
	Figures~\ref{fig:v11},~\ref{fig:sentinel},
	and~\ref{fig:expert} for V11, V13 SentinelNet, and V14
	ExpertNet respectively, with Figure~\ref{fig:eda}
	providing class distributions across splits for
	reference.

	
	\section{Results}
	\label{sec:results}
	
	\subsection{Flat Baseline}
	\label{sec:res_baseline}
	
	The flat baseline (V11) converges at epoch 9 of 20,
	with the combined selection criterion
	($0.6 \cdot \text{F1-Damage} +
	0.4 \cdot \text{F1-Growth}$) reaching 0.7654.
	F1-Damage --- reported as 3-class macro across Healthy,
	Drought, and Weeds --- is 0.777. Accuracy on the damage
	task is 0.803. F1-Growth is 0.748, with accuracy 0.785.
	Training curves are shown in Figure~\ref{fig:v11}:
	loss decreases smoothly through epoch 9 and the
	F1-Damage score plateaus around 0.77--0.78 for the
	remainder of training, indicating that the model
	extracts most available signal from the joint 3-class
	objective early and does not meaningfully improve
	thereafter. This plateau is
	consistent with the capacity limits of a flat
	architecture attempting to simultaneously model health
	status, damage type, and growth stage from a single
	shared representation without structural enforcement of
	the conditional label dependencies described in
	Section~\ref{sec:data} \citep{standley2020,
		caruana1997}.
	
	\subsection{Safety Performance}
	\label{sec:res_safety}
	
	The Sentinel (V13) reaches its best checkpoint at epoch
	4, with a Safety Score (defined in
	Section~\ref{sec:sentinel}) of 0.8532. The
	threshold at this checkpoint is $\tau = 0.56$. Training
	dynamics are shown in Figure~\ref{fig:sentinel}:
	Rec-Damaged remains above 0.95 at every epoch
	throughout the full 30-epoch training run, confirming
	that the safety constraint is satisfied at the level of
	checkpoint selection consistently and not only at the
	selected checkpoint \citep{cannon2002}. The threshold drifts from 0.56 at epoch 4
	to 0.19 at epoch 30, reflecting the model's increasing
	confidence in damage predictions as training progresses
	\citep{minderer2021}. The $\text{Rec-Damaged} \geq 0.95$
	floor reflects the operational requirement that at most
	5 missed damage cases per 100 truly damaged plots are
	tolerable, consistent with the irreversibility of claim
	denial in picture-based insurance
	\citep{barnett2007, ceballos2019}.

	At $\tau = 0.5$, the cascade achieves
	$\text{Rec-Damaged} = 0.974$ on the validation set. The
	test set contains approximately 2,616 truly damaged
	samples. At the flat baseline's Rec-Damaged of 0.943,
	approximately 148 damaged cases are missed. At the
	cascade's Rec-Damaged of 0.974, approximately 68 cases
	are missed --- a reduction of 80 missed damage cases,
	corresponding to a 3.1 percentage point improvement and
	roughly 54\% fewer missed damage cases in absolute
	terms. Experiment~1 (Section~\ref{sec:res_hypotheses})
	confirms this reduction is not explained by threshold
	calibration of the flat baseline alone.
	In an insurance-linked workflow \citep{barnett2007,
		jensen2016}, each missed damage case represents a
	farmer who sustained real losses and received no payout.
	That reduction is not a metric abstraction.
	
	Rec-Healthy at $\tau = 0.5$ is 0.424. This is low, and
	it should be stated plainly rather than explained away.
	The Sentinel is trained under a constraint that
	prioritizes damage recall over healthy precision ---
	the safety floor is $\text{Rec-Damaged} \geq 0.95$,
	not Rec-Healthy $\geq$ any bound beyond the minimum
	0.40 imposed during checkpoint selection
	\citep{elkan2001, cannon2002}. The model does exactly
	what it was trained to do: it aggressively routes
	toward damage detection, accepting a high false-alarm
	rate on healthy crops as the explicit operational cost
	of the safety constraint. At $\tau = 0.6$, Rec-Healthy
	improves to 0.650 with $\text{Rec-Damaged} = 0.929$,
	below the 0.95 target but within a range that may be
	acceptable in lower-risk deployment contexts
	\citep{barnett2008}. The $\tau$ parameter gives
	operators direct control over this trade-off without
	retraining \citep{geifman2017, bolukbasi2017}.

	The $\text{Rec-Damaged} \geq 0.95$ floor is one point
	on a continuous trade-off curve. Lower constraints
	reduce the safety floor but increase the pool of
	feasible checkpoints and the achievable Safety Score.
	Higher constraints reduce feasible checkpoints --- at
	$\text{Rec-Damaged} \geq 0.99$, very few epochs may
	satisfy the joint constraint, potentially yielding
	Safety Score $= 0$ for all epochs. The 0.95 value was
	selected as a conservative operational floor consistent
	with the irreversibility of false negatives in insurance
	assessment; operators with different risk tolerances
	may adjust this constraint without retraining by
	modifying the checkpoint selection criterion and
	rerunning checkpoint evaluation on locally
	representative validation data.

	\subsection{Expert Diagnostic Quality}
	\label{sec:res_expert}
	
	ExpertNet (V14; Section~\ref{sec:expert}) reaches its
	best checkpoint at epoch 7,
	with the composite selection score
	($0.7 \cdot \text{F1-Expert} +
	0.3 \cdot \text{Rec-Healthy}$) of 0.8278. F1-Expert
	--- macro F1 across Drought and Weeds, computed on
	truly damaged samples --- is 0.868. Per-class scores
	are F1-WED $= 0.944$ and F1-DGT $= 0.791$. Training
	dynamics are shown in Figure~\ref{fig:expert}: F1-WED
	converges rapidly and stabilizes above 0.94 by epoch 5;
	F1-DGT climbs more slowly and shows greater variance,
	consistent with the minority class status of Drought
	(20.1\% of damaged samples) and the 4:1 class
	imbalance \citep{lin2017, cui2019, he2009}. The
	composite HP-Score stabilizes around 0.82 from epoch 10
	onward.
	
	The Drought result requires direct engagement.
	F1-DGT $= 0.791$ means that on the test set's
	approximately 527 drought-damaged samples, roughly 110
	cases are misclassified --- either predicted as Weeds
	or, at the Sentinel stage, missed entirely before
	reaching the Expert. FocalLoss with $\gamma = 2.0$
	\citep{lin2017} partially addresses the 4:1
	Weeds/Drought imbalance by down-weighting easy majority
	examples, but does not resolve it. The gap between
	F1-WED $= 0.944$ and F1-DGT $= 0.791$ is not a
	rounding error --- it reflects a real representational
	challenge \citep{saleem2019, toda2019}. Drought stress
	and weed competition produce visually overlapping
	symptoms in maize at certain growth stages: leaf
	yellowing, stunted growth, and canopy thinning occur in
	both conditions \citep{coulibaly2019, sibiya2019}. At
	F1-DGT $= 0.791$, the system may not be sufficient for
	deployment settings that require treatment-level
	reliability, where distinguishing drought-appropriate
	responses from weed management interventions has direct
	agronomic consequences \citep{ceballos2019, barnett2007}.
	
	A note on comparability. V11's F1-Damage of 0.777 is a
	3-class macro score computed across Healthy, Drought,
	and Weeds on the full evaluation set. The cascade's
	F1-Expert of 0.868 is a 2-class macro score computed on
	truly damaged samples only --- healthy samples never
	reach the Expert and are excluded from this metric by
	design. These metrics measure different things and a
	direct numerical comparison between them is not
	methodologically valid \citep{davis2006, powers2011}.
	The cascade's higher expert-class score reflects in
	part the task decomposition: the Expert solves a
	simpler conditional problem rather than the harder
	joint problem the flat model faces \citep{standley2020}.
	This decomposition is the architectural contribution,
	not a confound to be corrected. It follows directly
	from the conditional label structure established in
	Section~\ref{sec:data}: damage type is only meaningful
	given damage presence, and the cascade enforces this
	structure at inference \citep{caruana1997, ruder2017}.
	A fair evaluation of whether the Expert's
	representational quality exceeds the flat model's on
	the same samples would require restricting the baseline
	evaluation to damaged samples --- an analysis not
	conducted here and noted as a methodological gap.
	Specifically, renormalizing V11's output probabilities
	over the damage classes $\{P(\text{DGT}|\mathbf{x}),
	P(\text{WED}|\mathbf{x})\}$ on the truly damaged subset
	would test whether the learned representation retains
	discriminative capacity when the healthy class is
	removed at inference time --- a test of representation
	reuse under evaluation alignment, not a test of
	equivalent training conditions: V11 was trained on all
	samples including healthy crops, while ExpertNet was
	trained exclusively on damaged samples with progressive
	upweighting.
	
	\subsection{Threshold Sweep}
	\label{sec:res_sweep}
	
\begin{figure*}[htbp]
		\centering
		\includegraphics[width=\textwidth]{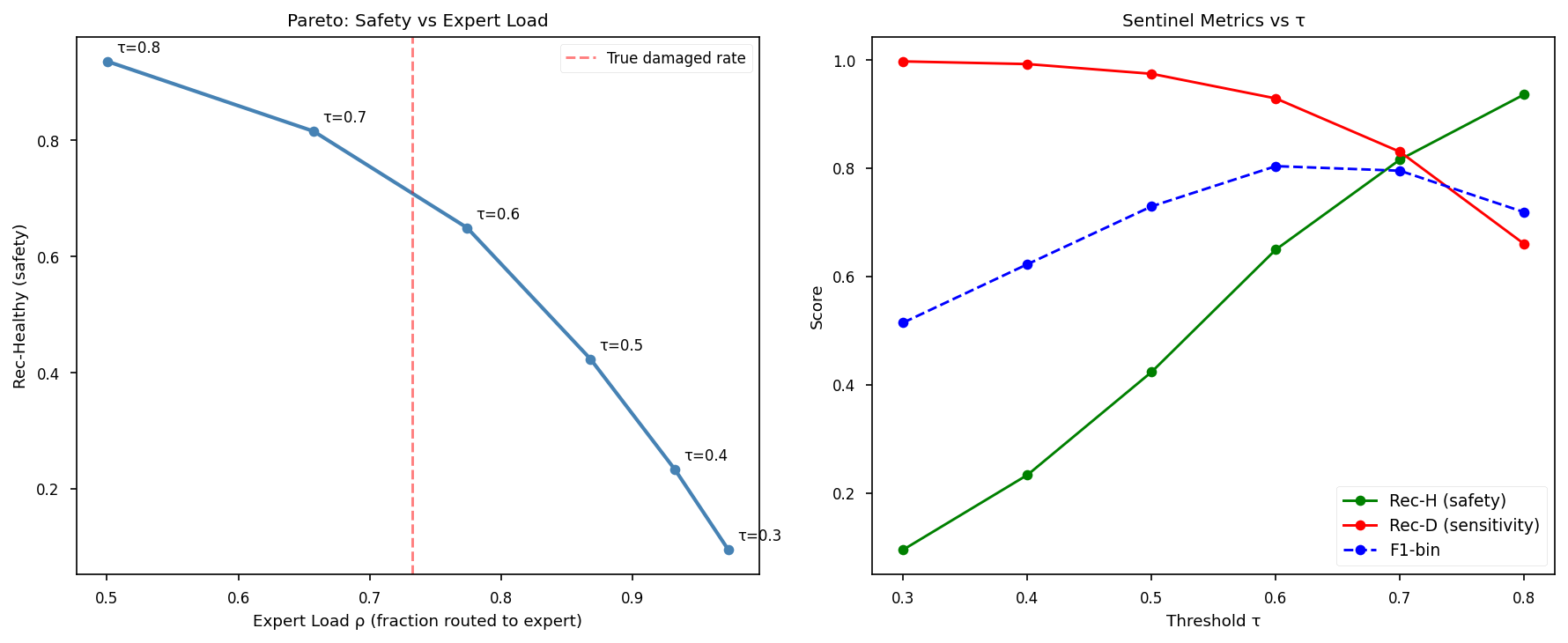}
		\caption{Cascade threshold sweep and Pareto frontier
			on the validation set ($n = 3{,}571$).
			\textbf{Left:} Pareto frontier between expert load
			$\rho$ (x-axis, fraction routed to ExpertNet) and
			Rec-Healthy (y-axis), with each $\tau$ value marked.
			The dashed vertical line indicates the true damaged
			rate (73.3\%). Operating points $\tau = 0.5$ and
			$\tau = 0.6$ lie in the region of meaningful
			safety-efficiency trade-off.
			\textbf{Right:} Rec-Damaged (red), Rec-Healthy
			(green), and F1-bin (blue dashed) as functions of
			$\tau.$ Rec-Damaged falls steeply above $\tau = 0.6$,
			reaching 0.660 at $\tau = 0.8$ --- below the
			threshold acceptable for insurance-linked deployment
			\citep{barnett2007, ceballos2019}.}
		\label{fig:threshold}
	\end{figure*}

	The cascade is evaluated across six operating points
	$\tau \in \{0.3, 0.4, 0.5, 0.6, 0.7, 0.8\}$ on the
	validation set ($n = 3{,}571$). Results are reported
	in Table~\ref{tab:sweep} and visualized in
	Figure~\ref{fig:threshold}. The left panel of
	Figure~\ref{fig:threshold} shows the Pareto frontier
	between expert load $\rho$ (x-axis) and Rec-Healthy
	(y-axis), with each $\tau$ value marked explicitly. The
	right panel shows Rec-Damaged, Rec-Healthy, and F1-bin
	as functions of $\tau$. Expert load $\rho$ serves as a
	proxy for the activation rate of the damage decision
	region, quantifying how input corruption or threshold
	variation shifts the effective routing boundary under
	a fixed model.
	
	Two properties of this sweep deserve explanation before
	the numbers are read. F1-Expert and F1-Growth are
	invariant to $\tau$ across all six operating points ---
	both remain at 0.868 and 0.828 respectively regardless
	of threshold setting. This is a property of the
	evaluation protocol, not of the architecture. ExpertNet
	is evaluated on the full truly-damaged set at every
	operating point, independent of what the Sentinel
	actually routes; growth prediction always comes from
	the Sentinel. In deployment,
	where only routed samples reach ExpertNet, both metrics
	would depend on routing composition and would vary with
	$\tau$ \citep{ovadia2019, taori2020}. Readers should
	interpret the $\tau$-invariant Expert and Growth scores
	as measures of intrinsic model quality, not as
	predictions of deployed system behavior.
	
	\begin{table*}[htbp]
		\centering
%
			\caption{Cascade threshold sweep on the validation
				set ($n = 3{,}571$) across $\tau \in \{0.3, \ldots,
				0.8\}$. Expert load $\rho$ denotes the fraction of
				samples routed to ExpertNet. F1-Expert and F1-Growth
				are $\tau$-invariant under this evaluation protocol
				(see Section~\ref{sec:eval}); both remain at 0.868
				and 0.828 respectively across all operating points
				because ExpertNet is evaluated on the full
				truly-damaged set independent of routing, and growth
				prediction always comes from the Sentinel.}
		\label{tab:sweep}
		\small
		\begin{tabular}{ccccccccc}
			\toprule
			$\tau$ & $\rho$ & \textbf{Expert} &
			\textbf{Rec-H} & \textbf{Rec-D} &
			\textbf{F1-bin} & \textbf{F1-Exp} &
			\textbf{F1-Gro} \\
			& & \textbf{Reduction} & & & & & \\
			\midrule
			0.3 & 0.973 & 2.7\%  & 0.095 & 0.997 &
			0.515 & 0.868 & 0.828 \\
			0.4 & 0.932 & 6.8\%  & 0.233 & 0.992 &
			0.623 & 0.868 & 0.828 \\
			0.5 & 0.868 & 13.2\% & 0.424 & 0.974 &
			0.730 & 0.868 & 0.828 \\
			0.6 & 0.774 & 22.6\% & 0.650 & 0.929 &
			0.804 & 0.868 & 0.828 \\
			0.7 & 0.657 & 34.3\% & 0.816 & 0.830 &
			0.795 & 0.868 & 0.828 \\
			0.8 & 0.500 & 50.0\% & 0.936 & 0.660 &
			0.719 & 0.868 & 0.828 \\
			\bottomrule
		\end{tabular}
	\end{table*}
	
	At $\tau = 0.3$, $\rho = 0.973$ --- nearly the entire
	dataset is routed to the Expert, and
	$\text{Rec-Damaged} = 0.997$ at the cost of
	$\text{Rec-Healthy} = 0.095$. The safety guarantee is
	maximized but the efficiency benefit is eliminated. At $\tau = 0.8$,
	$\rho = 0.500$ --- half the dataset bypasses the Expert
	--- but Rec-Damaged drops to 0.660, a miss rate that
	is too high for insurance-linked deployment
	\citep{barnett2007, jensen2016}. The operating point
	$\tau = 0.5$ provides $\text{Rec-Damaged} = 0.974$
	with a 13.2\% reduction in expert inference volume
	under clean input conditions, satisfying the safety
	constraint. This efficiency gain is conditional:
	it collapses to zero under severe input degradation
	as $\rho \to 1$ (Section~\ref{sec:disc_failure})
	\citep{bolukbasi2017, huang2017}. The operating point $\tau = 0.6$ provides
	the best F1-bin of 0.804 with a 22.6\% expert load
	reduction, at the cost of $\text{Rec-Damaged} = 0.929$
	--- below the 0.95 training target but potentially
	acceptable under constrained expert bandwidth where
	some safety margin can be traded for operational
	efficiency \citep{geifman2017}.
	Section~\ref{sec:disc_tau} discusses the deployment
	implications of this trade-off.
	
	
	\subsection{Robustness Under Distribution Shift}
	\label{sec:res_robust}
	
	This result demonstrates \emph{error containment via
	architectural decoupling} rather than general model
	robustness: because corruption at the Sentinel stage
	cannot propagate to ExpertNet's inputs, ExpertNet's
	input distribution does not change and its diagnostic
	outputs do not change.

\begin{figure*}[htbp]
		\centering
		\includegraphics[width=\textwidth]{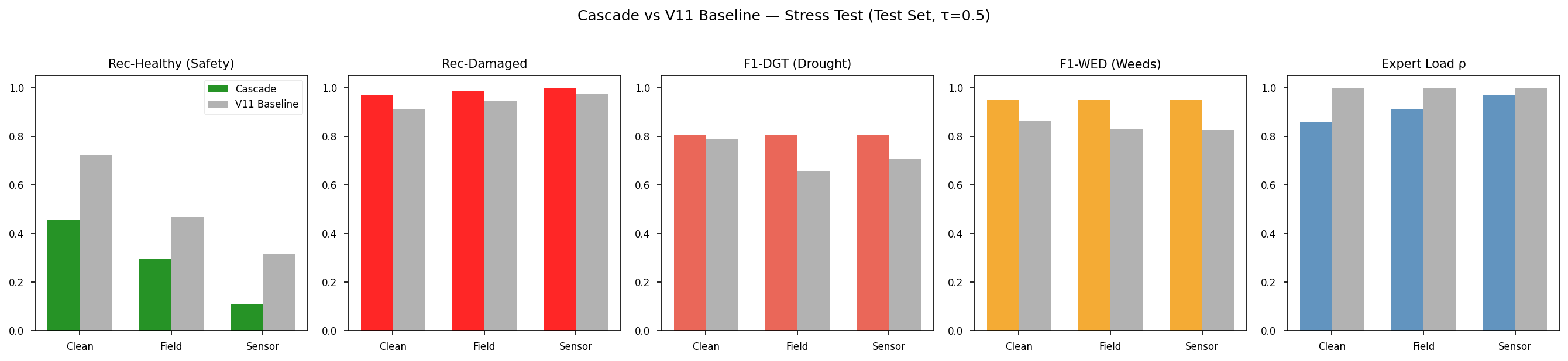}
		\caption{Cascade vs.\ V11 baseline robustness stress
			test on the test set ($n = 3{,}571$, $\tau = 0.5$). Five metrics shown across
			Clean, Field, and Sensor corruption conditions
			applied exclusively to Sentinel inputs (cascade) or
			uniformly to the single model (V11).
			\textbf{Rec-Healthy:} Cascade declines as $\rho$
			increases; V11 declines due to shared representation
			degradation \citep{hendrycks2019}.
			\textbf{Rec-Damaged:} Both systems maintain high
			recall; cascade increases to 0.997 under Sensor
			noise (safe failure mode).
			\textbf{F1-DGT:} Cascade stable at 0.805 across all
			conditions; V11 drops to 0.654 under Field ($-$17\%)
			and 0.708 under Sensor ($-$14\%) relative to Clean.
			\textbf{F1-WED:} Cascade stable at 0.949; V11
			degrades moderately \citep{geirhos2019}.
			\textbf{Expert Load $\rho$:} Cascade $\rho$ increases
			from 0.856 (Clean) to 0.968 (Sensor), reflecting
			the safe failure mode \citep{geifman2017}. V11
			$\rho = 1.0$ throughout (no routing mechanism).}
		\label{fig:stress}
	\end{figure*}

	The stress test is conducted on the test set
	($n = 3{,}571$) under three conditions applied
	exclusively to Sentinel inputs: Clean, Field
	(ColorJitter + GaussianBlur), and Sensor (additive
	Gaussian noise $\sigma = 0.08$). ExpertNet always
	receives clean images. This design reflects the
	deployment assumption that Sentinel inputs arrive from
	heterogeneous field devices while ExpertNet operates
	under controlled server-side preprocessing
	\citep{ceballos2019, waithaka2022}. It creates an
	intentionally asymmetric comparison --- cascade with
	partial corruption vs V11 with full corruption --- and
	should be read as testing whether architectural
	separation delivers practical value under the stated
	assumption, not as a neutral worst-case benchmark.
	Experiment~7 (Section~\ref{sec:res_hypotheses})
	examines the both-corrupted scenario. Results are
	shown in Figure~\ref{fig:stress} and
	Table~\ref{tab:stress}.
	
	Under all three conditions, ExpertNet's diagnostic metrics remain stable
	within measurement precision: F1-Expert $= 0.877$,
	F1-DGT $= 0.805$, F1-WED $= 0.949$ across Clean,
	Field, and Sensor conditions without variation. This
	stability is not a finding about model robustness in
	any general sense \citep{recht2019, geirhos2019} ---
	it is a direct consequence of the isolation property.
	Corruption applied to Sentinel inputs does not
	propagate to ExpertNet's inputs under the fixed
	preprocessing pipeline; therefore ExpertNet's input
	distribution does not change and its outputs do not
	change. The observed
	robustness predominantly arises from this
	architectural property rather than from
	representational invariance learned during training
	\citep{shorten2019}; whether limited intrinsic
	robustness also contributes under mild corruption is
	examined in Experiment~3
	(Section~\ref{sec:res_hypotheses}).
	
	What does change under corruption is $\rho$. Under
	Clean conditions, $\rho = 0.856$. Under Field
	corruption, $\rho$ increases to 0.912. Under Sensor
	noise, $\rho$ reaches 0.968 --- nearly the entire
	dataset is routed to the Expert, eliminating the
	efficiency benefit entirely. The Sentinel,
	uncertain under degraded inputs \citep{ovadia2019},
	routes more aggressively. This is a conservative
	failure mode --- the system errs toward expert
	consultation rather than toward confident
	misclassification \citep{geifman2017} --- but it
	has a practical cost: the efficiency gain of 13.2\%
	at Clean conditions ($\rho = 0.856$) disappears under
	Sensor noise ($\rho = 0.968$). Under constrained
	operational settings where expert inference capacity
	is limited \citep{ceballos2019, barnett2007}, Sensor-
	level input degradation effectively reverts the system
	to full Expert routing, and the cascade provides no
	throughput advantage over running ExpertNet on all
	inputs. Operators deploying in environments with
	frequent sensor degradation should treat $\rho$ as a
	function of input quality, not as a fixed operating
	parameter. Section~\ref{sec:disc_robust_caveat}
	discusses the architectural scope and limits of this
	guarantee.
	
	\begin{table*}[htbp]
		\centering
		\caption{Robustness stress test results on the test
			set ($n = 3{,}571$) under three corruption conditions
			applied exclusively to Sentinel inputs. ExpertNet
			always receives clean images. Expert-head metrics (F1-Expert, F1-DGT, F1-WED)
			remain stable within measurement precision across
			all conditions, consistent with the architectural
			isolation property (Section~\ref{sec:routing}). V11 has no architectural
			isolation; all metrics degrade simultaneously
			\citep{hendrycks2019}.}
		\label{tab:stress}
		\fontsize{7.5}{9.5}\selectfont
		\begin{tabular}{llccccccc}
			\toprule
			\textbf{Model} & \textbf{Condition} &
			\textbf{Rec-H} & \textbf{Rec-D} &
			$\rho$ & \textbf{F1-Exp} &
			\textbf{F1-DGT} & \textbf{F1-WED} &
			\textbf{F1-Gro} \\
			\midrule
			\multirow{3}{*}{Cascade}
			& Clean  & 0.455 & 0.969 & 0.856
			& 0.877 & 0.805 & 0.949 & 0.828 \\
			& Field  & 0.295 & 0.987 & 0.912
			& 0.877 & 0.805 & 0.949 & 0.811 \\
			& Sensor & 0.110 & 0.997 & 0.968
			& 0.877 & 0.805 & 0.949 & 0.775 \\
			\midrule
			\multirow{3}{*}{V11 Baseline}
			& Clean  & 0.722 & 0.912 & 1.000
			& --- & 0.787 & 0.863 & 0.729 \\
			& Field  & 0.468 & 0.944 & 1.000
			& --- & 0.654 & 0.827 & 0.644 \\
			& Sensor & 0.316 & 0.973 & 1.000
			& --- & 0.708 & 0.824 & 0.576 \\
			\bottomrule
		\end{tabular}
	\end{table*}
	
	The flat baseline shows no such protection. Under
	Field corruption, V11's F1-DGT drops from 0.787 to
	0.654 --- a 17\% relative degradation. F1-Growth drops
	from 0.729 to 0.644, a 12\% relative degradation.
	Under Sensor noise, Rec-Healthy falls from 0.722 to
	0.316 --- a 56\% relative drop --- and F1-Growth
	degrades to 0.576, 21\% below clean performance. These degradations occur
	simultaneously across all tasks because V11 has no
	architectural separation between corruption-sensitive
	and corruption-invariant components \citep{hendrycks2019,
		geirhos2019}. Corruption enters the shared
	representation and propagates everywhere. The
	comparison is shown directly in
	Figure~\ref{fig:stress}.
	
	The cascade's advantage over V11 under stress grows
	with corruption severity. Under Field conditions, the
	cascade maintains F1-DGT $= 0.805$ against V11's
	0.654 --- a 23\% relative advantage. Under Sensor
	noise, the cascade maintains F1-DGT $= 0.805$ against
	V11's 0.708 --- a 14\% relative advantage. For
	F1-Growth under Sensor noise, the cascade achieves
	0.775 against V11's 0.576 --- a 35\% relative
	advantage. These gaps do not exist under Clean
	conditions because architectural isolation only confers
	advantage when corruption is present
	\citep{taori2020, recht2019}. Under Clean conditions
	the cascade's Expert metrics slightly exceed V11's
	damage-type metrics, but the comparison remains
	indirect given the 3-class versus 2-class distinction
	established in Section~\ref{sec:res_expert}.
	
	One important caveat. The stress test applies
	corruption to Sentinel inputs only, reflecting a
	deployment assumption where ExpertNet operates under
	controlled server-side preprocessing while Sentinel
	inputs arrive from heterogeneous field devices
	\citep{ceballos2019, waithaka2022}. If in practice the
	preprocessing pipeline itself degrades --- due to
	transmission artifacts, lossy compression, or
	device-side preprocessing failures --- the isolation
	guarantee weakens \citep{ovadia2019}. The robustness
	demonstrated here is real under the stated assumptions;
	it should not be extrapolated to arbitrary pipeline
	configurations.
	
	
	\begin{figure*}[htbp]
		\centering
		\includegraphics[width=\textwidth]{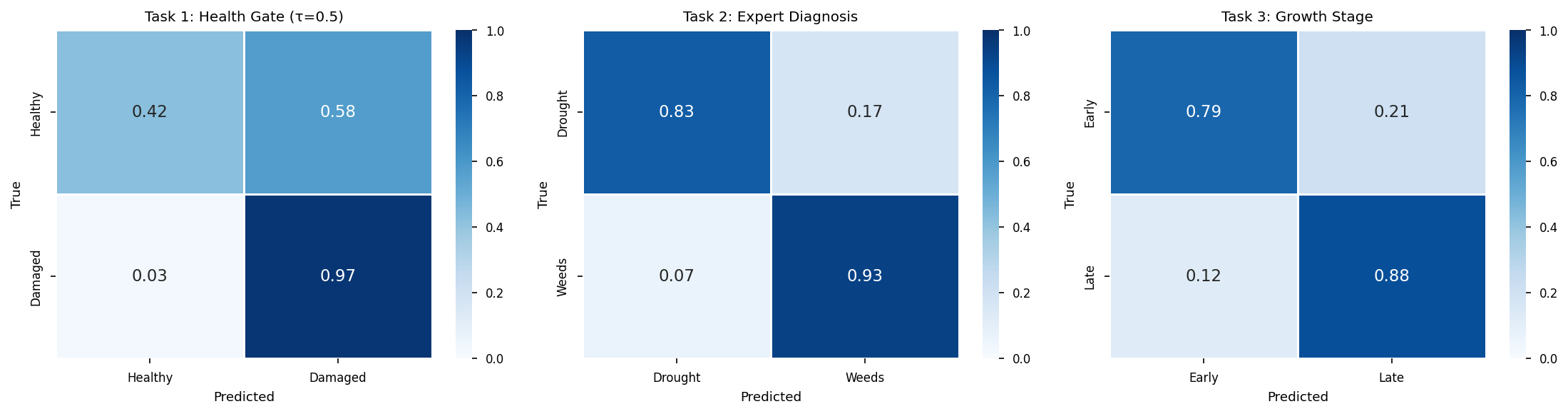}
		\caption{Normalized confusion matrices for all three
			tasks at $\tau = 0.5$ on the validation set.
			\textbf{Left (Task 1 --- Health Gate):} Sentinel
			binary triage. Rec-Damaged $= 0.97$; Rec-Healthy
			$= 0.42$, reflecting the asymmetric training
			objective \citep{elkan2001, cannon2002}.
			\textbf{Center (Task 2 --- Expert Diagnosis):}
			ExpertNet Drought/Weeds discrimination on truly
			damaged samples. F1-WED $= 0.944$; F1-DGT $= 0.791$,
			consistent with the 4:1 class imbalance and
			feature-space overlap between damage types
			\citep{lin2017, saleem2019}.
			\textbf{Right (Task 3 --- Growth Stage):} Sentinel
			auxiliary growth prediction. F1-Growth $= 0.828$,
			with Late growth (70.3\% of test set) correctly
			classified at 0.88 \citep{powers2011}.}
		\label{fig:confusion}
	\end{figure*}
	
	
	\subsection{Task Decomposition and the Comparison
		Question}
	\label{sec:res_decomp}
	
	The flat baseline's F1-Damage of 0.777 is a 3-class
	macro score computed over Healthy, Drought, and Weeds
	simultaneously \citep{powers2011}. The cascade's
	F1-Expert of 0.868 is a 2-class macro score computed on
	truly damaged samples only --- healthy samples never
	reach the Expert and are excluded from this metric by
	design. These metrics
	measure different things and a direct numerical
	comparison between them is not methodologically valid
	\citep{davis2006}.
	
	The cascade achieves higher expert-class accuracy
	partly because it exploits conditional label structure
	to reduce hypothesis complexity, trading a single
	harder joint problem for two simpler conditional ones
	\citep{standley2020, caruana1997}. This reduction is
	not a methodological convenience --- it is the
	architectural expression of the conditional label
	structure documented in Section~\ref{sec:data}:
	damage type is only meaningful given damage presence,
	and the cascade enforces this dependency at the system
	level. Both sets of numbers are reported transparently
	in Tables~\ref{tab:sweep} and~\ref{tab:stress}.

	The comparison between ExpertNet ($\text{F1} = 0.868$)
	and V11 ($\text{F1} = 0.777$) is not directly
	commensurable, as ExpertNet operates on a conditional
	binary task while V11 performs joint three-class
	classification. Under evaluation alignment
	(Experiment~2, Section~\ref{sec:res_hypotheses}),
	this gap reduces to $+0.008$ in F1-macro --- a
	difference too small to constitute a meaningful
	representational advantage on a single test set.
	The primary contribution of the cascade is therefore
	not improved representation learning, but the
	enforcement of safety constraints through structured
	inference at comparable diagnostic quality. Task decomposition is
	the contribution; the performance difference is
	its consequence.

	\subsection{Hypothesis Testing Experiments}
	\label{sec:res_hypotheses}

	Four targeted experiments evaluate the competing
	explanations for cascade behavior. Results are reported
	below; Section~\ref{sec:discussion} synthesizes the
	verdicts.

	\noindent\textbf{Experiment~1 --- Decision Boundary
		Hypothesis (H1).}
	H1 asks whether threshold calibration of V11 alone
	replicates cascade performance. On the hierarchical
	validation set, the lowest $\tau$ satisfying
	$\text{Rec-D} \geq 0.95$ is $\tau = 0.30$.
	Table~\ref{tab:exp1} reports test-set performance at
	this operating point. The safety constraint is met
	under all three conditions ($\text{Rec-D} \geq 0.965$),
	but at the cost of routing 80.8--95.6\% of samples to
	Expert, eliminating most of the efficiency benefit.
	Under Field corruption, F1-DGT falls from 0.877 to
	0.734 ($-16\%$) because the flat model's shared
	representation propagates corruption to all outputs
	without isolation. \textbf{H1 is rejected.}

	\begin{table}[htbp]
		\centering
		\caption{V11 with threshold-calibrated routing
			($\tau = 0.30$, selected on validation) evaluated on
			the test set ($n = 3{,}571$) under three corruption
			conditions (Experiment~1). The safety constraint is
			met but efficiency is eliminated; F1-DGT degrades
			under corruption due to shared-representation
			propagation.}
		\label{tab:exp1}
		\small
		\begin{tabular}{lcccccc}
			\toprule
			\textbf{Condition} & \textbf{Rec-D} &
			\textbf{Rec-H} & \textbf{F1-DGT} &
			\textbf{F1-WED} & \textbf{F1-Gro} & $\rho$ \\
			\midrule
			Clean  & 0.965 & 0.622 & 0.877 & 0.971 & 0.898 & 0.808 \\
			Field  & 0.987 & 0.289 & 0.734 & 0.948 & 0.853 & 0.913 \\
			Sensor & 0.995 & 0.150 & 0.802 & 0.952 & 0.852 & 0.956 \\
			\bottomrule
		\end{tabular}
	\end{table}

	\noindent\textbf{Experiment~2 --- Data Distribution
		Hypothesis (H2).}
	H2 asks whether V11's apparent diagnostic disadvantage
	survives evaluation alignment. When V11 is restricted
	to the truly damaged test subset ($n = 2{,}616$) and
	its 3-class output is renormalized to a 2-class
	(DGT/WED) probability, F1-macro $= 0.924$ against
	ExpertNet's $0.932$ --- a gap of only $+0.008$
	(Table~\ref{tab:exp2}). Evaluation mismatch between
	a 3-class model and a 2-class intrinsic evaluation
	accounts for most of the apparent advantage.
	\textbf{H2 is supported:} the cascade's diagnostic
	advantage is modest in representational terms;
	architectural isolation is a key mechanism enabling
	the cascade's structural advantage, not representational
	superiority.

	\begin{table}[htbp]
		\centering
		\caption{V11 (renormalized outputs, damaged subset)
			vs.\ ExpertNet on identical samples ($n = 2{,}616$;
			Experiment~2). The representational gap is 0.008
			F1-macro; evaluation mismatch accounts for the
			remainder.}
		\label{tab:exp2}
		\small
		\begin{tabular}{lccc}
			\toprule
			\textbf{Model} & \textbf{F1-DGT} &
			\textbf{F1-WED} & \textbf{F1-macro} \\
			\midrule
			V11 (renorm., damaged subset) & 0.877 & 0.971 & 0.924 \\
			ExpertNet (cascade, $\tau=0.5$) & 0.893 & 0.971 & 0.932 \\
			\midrule
			\emph{Gap} & \emph{+0.016} & \emph{0.000} & \emph{+0.008} \\
			\bottomrule
		\end{tabular}
	\end{table}

	\noindent\textbf{Experiment~3 --- Input Isolation
		Hypothesis (H3).}
	H3 asks whether ExpertNet's robustness arises from
	structural isolation or from intrinsic learned
	invariance. Corruption is applied directly to
	ExpertNet's $256 \times 256$ inputs on all $2{,}616$
	truly damaged test images, bypassing SentinelNet
	entirely. Table~\ref{tab:exp3} shows that under Sensor
	noise, F1-Expert falls from 0.932 to 0.745 and F1-DGT
	collapses from 0.893 to 0.570 ($-36\%$ relative).
	The stability in Table~\ref{tab:stress} is
	predominantly attributable to ExpertNet not receiving
	corrupted inputs under cascade operation; the mild
	Field degradation ($-1.7\%$ F1-Expert) suggests
	limited intrinsic resilience under weak perturbations.
	\textbf{H3 is strongly supported:} robustness is
	predominantly architectural rather than a property of
	the learned representation, though minor intrinsic
	robustness cannot be ruled out under mild corruption.

	\begin{table}[htbp]
		\centering
		\caption{ExpertNet evaluated with corruption applied
			directly to its inputs (SentinelNet bypassed;
			Experiment~3). Contrast with Table~\ref{tab:stress}
			where corruption applies to Sentinel inputs only.
			F1-DGT collapses $-36\%$ under Sensor noise,
			confirming that isolation drives the robustness.}
		\label{tab:exp3}
		\small
		\begin{tabular}{lccc}
			\toprule
			\textbf{Condition} & \textbf{F1-Expert} &
			\textbf{F1-DGT} & \textbf{F1-WED} \\
			\midrule
			Clean  & 0.932 & 0.893 & 0.971 \\
			Field  & 0.916 & 0.868 & 0.963 \\
			Sensor & 0.745 & 0.570 & 0.920 \\
			\bottomrule
		\end{tabular}
	\end{table}

	\noindent\textbf{Experiment~4 --- Control Parameter
		Hypothesis (H4).}
	H4 asks whether $\tau$ transfers across environments
	without local recalibration. Table~\ref{tab:exp4}
	reports Rec-D and $\rho$ on validation and test splits
	for six operating points. The maximum Rec-D gap is
	$\Delta = 0.013$ at $\tau = 0.7$; the maximum $\rho$
	gap is $\Delta = 0.011$ at the same point. Both remain
	below 0.02 throughout the sweep. \textbf{H4 is
	supported:} $\tau$ transfers reliably across held-out
	splits of the same distribution; practitioners may
	calibrate on a validation set and deploy without
	re-tuning under matched conditions. Cross-distribution
	transfer --- different geography, crop type, or imaging
	device --- would require re-calibration and is not
	tested here.

	\begin{table}[htbp]
		\centering
		\caption{$\tau$ transfer: validation vs.\ test
			($n = 3{,}571$ each; Experiment~4). $\Delta$
			denotes the absolute split difference. Maximum
			$\Delta\,\text{Rec-D} = 0.013$ at $\tau = 0.7$;
			$\tau$ calibrates stably within same-distribution
			splits.}
		\label{tab:exp4}
		\small
		\begin{tabular}{ccccccc}
			\toprule
			$\tau$ & \textbf{Rec-D val} & \textbf{Rec-D test} &
			$\Delta$ & $\rho$ \textbf{val} &
			$\rho$ \textbf{test} & $\Delta$ \\
			\midrule
			0.3 & 1.000 & 0.999 & 0.001 & 0.969 & 0.970 & 0.000 \\
			0.4 & 0.995 & 0.996 & 0.001 & 0.928 & 0.931 & 0.004 \\
			0.5 & 0.977 & 0.978 & 0.000 & 0.864 & 0.864 & 0.000 \\
			0.6 & 0.938 & 0.932 & 0.006 & 0.771 & 0.765 & 0.006 \\
			0.7 & 0.851 & 0.837 & \textbf{0.013} & 0.659 & 0.648 & \textbf{0.011} \\
			0.8 & 0.680 & 0.673 & 0.006 & 0.508 & 0.504 & 0.004 \\
			\bottomrule
		\end{tabular}
	\end{table}

	\noindent\textbf{Experiment~5 --- Oracle Routing Upper
		Bound.}
	The oracle routes all truly damaged samples directly to
	ExpertNet and predicts Healthy for all healthy samples,
	eliminating all routing error. Table~\ref{tab:exp5}
	shows that cascade ($\tau = 0.5$) and oracle produce
	identical F1-Expert, F1-DGT, and F1-WED (gap $=
	0.000$). This arises from the intrinsic evaluation
	protocol: ExpertNet is evaluated on the full damaged
	subset regardless of routing, so routing quality is not
	captured by Expert metrics alone. The oracle analysis
	confirms that, under the intrinsic evaluation protocol,
	\emph{classification} performance defines the
	measurable system ceiling: improving ExpertNet's
	diagnostic accuracy yields greater gain than improving
	SentinelNet's routing precision \emph{as measured here}.
	In deployment, where routing errors directly determine
	which samples reach ExpertNet, routing quality would
	also surface in end-to-end metrics.

	\begin{table}[htbp]
		\centering
		\caption{Oracle routing upper bound vs.\ cascade at
			$\tau = 0.5$ on the test set (Experiment~5). Under
			the intrinsic evaluation protocol, oracle and cascade
			produce identical Expert metrics (gap $= 0.000$).
			The oracle is an analytical bound, not a deployable
			configuration.}
		\label{tab:exp5}
		\small
		\begin{tabular}{lcccccc}
			\toprule
			\textbf{Routing} & \textbf{Rec-D} &
			\textbf{Rec-H} & $\rho$ &
			\textbf{F1-Exp} & \textbf{F1-DGT} &
			\textbf{F1-WED} \\
			\midrule
			Oracle (perfect) & 1.000 & 1.000 & 0.733
			& 0.932 & 0.893 & 0.971 \\
			Cascade ($\tau=0.5$) & 0.978 & 0.449 & 0.864
			& 0.932 & 0.893 & 0.971 \\
			\bottomrule
		\end{tabular}
	\end{table}

	\noindent\textbf{Experiment~6 --- End-to-End Routing
		Bias Analysis.}
	The intrinsic evaluation protocol (Experiments 1--5)
	evaluates ExpertNet on the full truly-damaged subset
	regardless of routing. Experiment~6 closes this gap
	by running the cascade end-to-end: SentinelNet routes
	each sample, and ExpertNet processes only what it
	actually receives. Table~\ref{tab:exp6} reports
	routing composition and performance at five operating
	points.

	At $\tau = 0.5$, 526 truly healthy samples (55.1\%
	of all healthy crops) are routed to ExpertNet ---
	all of them receiving a false damage diagnosis, since
	ExpertNet has no Healthy class. Of these, 90\% are
	misclassified as WED (weed presence) and 10\% as DGT
	(drought), a systematic bias consistent with healthy
	crops sharing more visual features with the
	weed-affected class. The protocol gap between
	intrinsic and TP-routed Expert F1 is $-0.0004$ (below
	measurement precision), confirming that the
	intrinsic evaluation protocol is not artificially
	inflated relative to correctly-routed performance.
	The end-to-end 3-class F1-macro at $\tau = 0.5$ is
	0.767, against the intrinsic F1-Expert of 0.932 ---
	the gap reflects false positive routing, not
	ExpertNet's classification quality.

	\begin{table}[htbp]
		\centering
		\caption{End-to-end routing evaluation across five
			operating points (Experiment~6). TP, FP, FN, TN
			denote routing outcomes. FP-rate-H: fraction of
			healthy crops misrouted to Expert; all receive a
			false damage label. Protocol gap: intrinsic
			F1-Expert minus TP-routed F1-Expert; near-zero
			confirms the intrinsic protocol is accurate.
			F1-E2E: true 3-class macro F1 (Healthy/DGT/WED).}
		\label{tab:exp6}
		\fontsize{7.5}{9.5}\selectfont
		\begin{tabular}{ccccccccc}
			\toprule
			$\tau$ & TP & FP & FN & TN &
			\textbf{FP-rate-H} &
			\textbf{Gap} &
			\textbf{F1-E2E} \\
			& & & & & (\% healthy) & (intr$-$TP) & (3-class) \\
			\midrule
			0.3 & 2613 & 849 & 3 & 106 & 89\% & +0.000 & 0.618 \\
			0.4 & 2606 & 720 & 10 & 235 & 75\% & $-$0.000 & 0.690 \\
			0.5 & 2558 & 526 & 58 & 429 & 55\% & $-$0.000 & 0.767 \\
			0.6 & 2438 & 293 & 178 & 662 & 31\% & +0.000 & 0.821 \\
			0.7 & 2190 & 124 & 426 & 831 & 13\% & $-$0.001 & 0.811 \\
			\bottomrule
		\end{tabular}
	\end{table}

	This result has a design implication: to reduce false
	damage diagnoses for healthy crops, $\tau$ should be
	raised above 0.5. At $\tau = 0.6$, the FP route rate
	drops from 55\% to 31\% and end-to-end F1-macro
	improves from 0.767 to 0.821 --- but at the cost of
	Rec-D falling from 0.978 to 0.932, below the 0.95
	safety floor. The operating point is a genuine
	trade-off between missed damage recall and healthy
	crop misdiagnosis.

	\noindent\textbf{Experiment~7 --- Both-Corrupted
		Stress Test.}
	To close the comparison asymmetry in
	Table~\ref{tab:stress} (cascade partially corrupted
	vs V11 fully corrupted), Experiment~7 applies the
	same corruption to both Sentinel (224$\times$224)
	and Expert (256$\times$256) inputs simultaneously.
	Table~\ref{tab:exp7} reports all seven scenarios.
	Three findings emerge.

	First, the isolation boundary remains effective under
	combined corruption. Both-Field and Expert-only-Field
	produce nearly identical Expert metrics (F1-Exp 0.914
	vs 0.913, F1-DGT 0.865 vs 0.864). Both-Sensor and
	Expert-only-Sensor are similarly indistinguishable
	(F1-Exp 0.745 vs 0.744, F1-DGT 0.569 vs 0.568).
	Sentinel corruption adds no marginal Expert
	degradation because Sentinel and Expert inputs are
	independently preprocessed: even when both stages
	are corrupted, the Expert's performance is determined
	solely by what reaches its input, not by what
	happened upstream.

	Second, pipeline failures are additive, not
	multiplicative. The interaction term is $+0.001$
	for both Field and Sensor conditions --- below
	measurement precision. There is no compound failure
	mode: corrupting Sentinel and Expert simultaneously
	produces exactly the sum of corrupting each
	independently.

	Third, when the pipeline assumption fully fails
	(Both Sensor), the cascade provides no robustness
	advantage: F1-DGT collapses to 0.569 ($-$36\%) and
	efficiency is eliminated ($\rho = 0.923$). The
	system's safety properties require maintaining
	ExpertNet input quality. If that condition holds,
	the isolation delivers the robustness shown in
	Table~\ref{tab:stress}; if it fails, performance
	matches the Expert-only corrupted scenario.

	\begin{table*}[htbp]
		\centering
		\caption{Both-corrupted stress test at $\tau = 0.5$
			(Experiment~7). Seven scenarios compare Sentinel-only,
			Expert-only, and simultaneous corruption. Interaction
			terms (Both minus additive sum) are $+0.001$ for Field
			and Sensor, confirming that failures are additive and
			the isolation boundary contains Sentinel-stage
			corruption even under simultaneous Expert corruption.
			All metrics on test set ($n = 3{,}571$) using the
			intrinsic evaluation protocol.}
		\label{tab:exp7}
		\fontsize{7.5}{9.5}\selectfont
		\begin{tabular}{llcccccc}
			\toprule
			\textbf{Sentinel} & \textbf{Expert} &
			\textbf{Rec-D} & \textbf{Rec-H} & $\rho$ &
			\textbf{F1-Exp} & \textbf{F1-DGT} & \textbf{F1-WED} \\
			\midrule
			Clean  & Clean  & 0.978 & 0.449 & 0.864
			       & 0.932  & 0.893 & 0.971 \\
			\midrule
			Field  & Clean  & 0.989 & 0.269 & 0.920
			       & 0.932  & 0.893 & 0.971 \\
			Clean  & Field  & 0.978 & 0.449 & 0.864
			       & 0.913  & 0.864 & 0.962 \\
			\textbf{Field}  & \textbf{Field}  & 0.991 & 0.276 & 0.919
			       & \textbf{0.914}  & \textbf{0.865} & \textbf{0.962} \\
			\midrule
			Sensor & Clean  & 0.976 & 0.212 & 0.926
			       & 0.932  & 0.893 & 0.971 \\
			Clean  & Sensor & 0.978 & 0.449 & 0.864
			       & 0.744  & 0.568 & 0.920 \\
			\textbf{Sensor} & \textbf{Sensor} & 0.975 & 0.220 & 0.923
			       & \textbf{0.745}  & \textbf{0.569} & \textbf{0.920} \\
			\bottomrule
		\end{tabular}
	\end{table*}

	
	\section{Discussion}
	\label{sec:discussion}
	
	The results reported in Section~\ref{sec:results} are
	consistent with the paper's organizing hypothesis, but
	consistency is not the same as confirmation, and the
	discussion that follows tries to hold that distinction
	seriously \citep{ovadia2019, taori2020}.

	Four competing explanations for the observed cascade
	behavior deserve explicit evaluation. The
	\emph{decision boundary hypothesis} asks whether
	threshold calibration of a flat model alone replicates
	cascade behavior. The \emph{data distribution
	hypothesis} asks whether the diagnostic advantage
	survives evaluation alignment --- that is, whether
	V11's representational quality on the damaged subset
	is genuinely inferior or whether the gap reflects
	evaluation mismatch between a 3-class model and a
	2-class intrinsic evaluation. The \emph{input
	isolation hypothesis} asks whether robustness arises
	from structurally decoupling the diagnostic model from
	corrupted upstream inputs, rather than from learned
	representational invariance. The \emph{control
	parameter hypothesis} asks whether $\tau$ transfers
	across environments or requires local recalibration.
	Section~\ref{sec:res_hypotheses} reports five targeted
	experiments that evaluate each hypothesis directly.
	H1 is rejected: threshold calibration of a flat model
	satisfies the safety constraint only at the cost of
	eliminating efficiency, and the flat representation
	still propagates corruption to diagnostic outputs.
	H2 is supported: under fair evaluation on identical
	samples, the representational gap shrinks to 0.008
	F1-macro, confirming that evaluation mismatch drove
	most of the apparent advantage. H3 is strongly
	supported: ExpertNet degrades severely when directly
	corrupted (F1-DGT $-36\%$ under Sensor noise),
	confirming that isolation is the predominant source of
	robustness; limited intrinsic resilience may persist
	under mild Field corruption ($-1.7\%$ F1-Expert). H4
	is supported: the maximum Rec-D transfer gap is 0.013
	across same-distribution splits, confirming stable
	calibration within distribution; cross-distribution
	transfer is not tested. The oracle analysis confirms
	that, under the intrinsic evaluation protocol,
	classification performance defines the measurable
	system ceiling.

	Taken together, the five experiments support a
	consistent framing: system performance is governed by
	a joint objective over routing accuracy
	($\text{Rec-D}$, $\rho$) and conditional
	classification performance (F1-DGT, F1-WED) that
	cannot be reduced to a single classifier metric.
	The two objectives are coupled in deployment but
	partially decoupled by the intrinsic evaluation
	protocol, which is why the oracle gap is zero while
	routing errors remain present.

	\subsection{\texorpdfstring{$\tau$}{tau} as an
		Operational Decision}
	\label{sec:disc_tau}
	
	The threshold $\tau$ is not a tuning parameter in the
	conventional sense. It does not affect model weights,
	training dynamics, or ExpertNet's learned parameters
	or its intrinsic performance under a fixed evaluation
	distribution. As shown in Table~\ref{tab:sweep} and 
	Figure~\ref{fig:threshold}, F1-Expert remains fixed
	at 0.868 and F1-Growth at 0.828 across all six
	operating points $\tau \in \{0.3, 0.4, 0.5, 0.6, 0.7,
	0.8\}$. What $\tau$ controls is the composition of the
	routing decision: how much of the input stream is sent
	to the Expert, and at what cost to Rec-Damaged
	\citep{elkan2001, cannon2002}. This is an operational
	decision that belongs to the deploying organization,
	not a modeling decision that belongs to the training
	process \citep{ceballos2019, barnett2007}.
	
	Two operating points emerge from the data as natural
	anchors. At $\tau = 0.5$, $\text{Rec-Damaged} = 0.974$
	and $\rho = 0.868$, meaning 13.2\% of samples bypass
	the Expert entirely (Table~\ref{tab:sweep}). This is
	the safety-first configuration: the recall constraint
	is satisfied with margin, at the cost of routing the
	majority of the input stream to ExpertNet. At $\tau =
	0.6$, $\text{Rec-Damaged} = 0.929$ and $\rho = 0.774$,
	achieving 22.6\% expert load reduction with
	F1-bin $= 0.804$ --- the best binary classification
	performance across the sweep \citep{bolukbasi2017,
		huang2017}. Whether $\tau = 0.5$ or $\tau = 0.6$ is
	appropriate depends on the operator's tolerance for
	missed damage cases relative to Expert model capacity
	--- a decision the system supports but cannot make
	\citep{jensen2016, barnett2008}.

	To make the trade-off concrete: at $\tau = 0.5$, the
	cascade routes 86.8\% of submitted images to ExpertNet,
	corresponding to an expert load reduction of 13.2\%
	compared to routing all images. For a deployment
	processing 10,000 images per season, this corresponds
	to approximately 1,320 fewer expert reviews --- a
	direct reduction in operational cost without
	compromising the $\text{Rec-Damaged} \geq 0.95$ safety
	floor. At $\tau = 0.6$, the reduction increases to
	22.6\%, or approximately 2,260 fewer reviews per
	10,000 images, at the cost of
	$\text{Rec-Damaged} = 0.929$. These figures are
	deployment-scale quantities, not percentage points:
	they translate directly into agronomist workload and
	insurance processing capacity for the operator.

	At $\tau = 0.5$, 57.6\% of truly healthy plots
	are routed to ExpertNet as potentially damaged,
	implying that an insurance organization processing
	10,000 plots per season would generate approximately
	2,723 unnecessary expert reviews alongside the
	1,320 avoided through expert load reduction.
	Increasing the threshold to $\tau = 0.6$ improves
	specificity ($\text{Rec-Healthy} = 0.650$), reducing
	unnecessary reviews to 35\% of healthy plots.
	More generally, given a relative cost ratio between
	false negatives and false positives, the optimal
	operating threshold $\tau$ can be selected by
	minimizing expected decision cost over a validation
	set, enabling deployment-specific calibration
	according to institutional risk preferences
	without retraining.

	This control is meaningful under stable
	deployment conditions; it requires
	revalidation on locally representative data
	as Sentinel probability calibration drifts,
	and the operating characteristics associated
	with a fixed $\tau$ should not be assumed to
	transfer across geographies or sensor
	environments without empirical verification
	(Section~\ref{sec:limitations}).
	
	At the extremes, the Pareto frontier becomes
	uninformative. At $\tau = 0.3$, $\rho = 0.973$ and
	$\text{Rec-Damaged} = 0.997$ --- virtually no damage
	is missed, but the Expert processes nearly everything,
	eliminating the cascade's efficiency rationale
	\citep{panda2016}. At $\tau = 0.8$,
	$\rho = 0.500$ and $\text{Rec-Damaged} = 0.660$ ---
	half the input stream bypasses the Expert, but the
	probability of missing real damage is unacceptably high
	for an insurance-linked workflow \citep{barnett2007,
		ceballos2019}. The usable operating range is roughly
	$\tau \in [0.5, 0.7]$, where the Pareto frontier
	offers meaningful trade-offs between safety and
	efficiency, as visualized in
	Figure~\ref{fig:threshold}. While thresholding can be
	applied to flat classifiers \citep{menardi2014,
		wallace2011}, the resulting trade-off is not
	structurally decoupled from representation learning
	and, as observed in Section~\ref{sec:res_robust},
	degrades more sharply under distribution shift in this
	setting \citep{hendrycks2019, taori2020}.
	
	\subsection{Robustness: What the Architecture Provides
		and What It Requires}
	\label{sec:disc_robust_caveat}

	ExpertNet does not exhibit robustness to corrupted
	inputs when evaluated in isolation
	(Experiment~3): under direct Sensor noise,
	F1-DGT collapses from 0.893 to 0.570, a 36\%
	relative degradation. The stability reported in
	Table~\ref{tab:stress} arises entirely from the
	preprocessing pipeline delivering clean images to
	ExpertNet regardless of Sentinel-input corruption.
	Robustness in this system therefore arises from
	architectural isolation and error containment, not
	from learned invariance within the model itself.

	Experiment~7 (Section~\ref{sec:res_hypotheses})
	resolves the asymmetric-comparison concern in
	Table~\ref{tab:stress}: when corruption is applied
	to both stages simultaneously, failures are additive
	($\text{interaction} = +0.001$, below measurement
	precision) and the isolation boundary remains
	effective. Both-corrupted Field and Sensor scenarios
	are empirically indistinguishable from their
	Expert-only counterparts, because Sentinel and
	Expert inputs are independently preprocessed. The
	cascade does not amplify simultaneous failures.

	This result sharpens the precision of the robustness
	claim. The architecture provides one guarantee: if
	ExpertNet's input quality is maintained, ExpertNet's
	outputs are unaffected by what happens at the
	Sentinel stage. Experiment~7 confirms this guarantee
	holds even when Sentinel is simultaneously corrupted.
	What the architecture cannot provide is robustness
	against ExpertNet input corruption: under Sensor
	noise applied to ExpertNet's inputs, F1-DGT
	collapses from 0.893 to 0.569 ($-$36\%) whether or
	not Sentinel is also corrupted. The system's
	robustness depends entirely on maintaining ExpertNet
	input quality at deployment --- a pipeline
	requirement, not a model property.

	Practically, this means the robustness guarantee in
	Table~\ref{tab:stress} should be read as conditional:
	it holds under the deployment assumption that
	ExpertNet receives server-side preprocessed images
	while Sentinel inputs arrive from heterogeneous field
	devices. If transmission artifacts, lossy compression,
	or device-side preprocessing failures reach ExpertNet
	inputs, the system degrades to the Expert-only
	corrupted scenario shown in
	Table~\ref{tab:exp7} --- a condition operators
	should monitor and guard against by design.

	\subsection{The Safe Failure Mode and Its Cost}
	\label{sec:disc_failure}
	
	The most practically significant finding in the
	robustness stress test (Table~\ref{tab:stress};
	Figure~\ref{fig:stress}) is not that the cascade
	outperforms V11 under corruption --- though it does,
	substantially --- but that its failure mode is
	structurally safe \citep{geifman2017, cannon2002}. In
	effect, the cascade converts input uncertainty into
	routing escalation rather than prediction error
	\citep{ovadia2019}. Under sensor noise ($\sigma = 0.08$
	additive Gaussian, applied post-normalization),
	Rec-Healthy drops to 0.110 and $\rho$ increases to
	0.968 at $\tau = 0.5$. The system is routing nearly
	everything to ExpertNet. This is not good performance.
	But it is a safe failure: the Sentinel, uncertain under
	degraded inputs \citep{minderer2021}, escalates rather
	than commits. Rec-Damaged increases to 0.997 under
	sensor noise because the Sentinel's bias toward routing
	removes the possibility of a false Healthy prediction
	on damaged crops. This
	safety guarantee, however, is contingent on ExpertNet
	maintaining its diagnostic reliability under the same
	shift conditions --- a property that holds here because
	the Expert always receives clean images, but which
	should not be assumed to generalize to pipeline
	configurations where Expert inputs are also corrupted
	\citep{ovadia2019, taori2020}.
	
	The cost of this safe failure mode is total elimination
	of the cascade's efficiency benefit. At $\rho = 0.968$,
	the Expert processes 96.8\% of the input stream ---
	functionally equivalent in routing volume to running
	ExpertNet on everything, though not necessarily
	identical in computational overhead due to pipeline
	structure \citep{bolukbasi2017, panda2016}. The
	cascade's 13--22\% expert load reduction, documented
	under clean conditions, disappears entirely under
	severe sensor degradation (Table~\ref{tab:stress}).
	Operators should plan for this: a system that claims
	computational efficiency under field conditions must
	account for the full distribution of sensor quality it
	will encounter \citep{ceballos2019, waithaka2022}, not
	just the clean-condition baseline.

	The efficiency benefit observed under clean
	conditions --- a 13.2\% reduction in expert reviews
	at $\tau = 0.5$ --- collapses under severe input
	degradation ($\rho = 0.968$ under Sensor noise),
	where most samples are escalated to ExpertNet.
	This behavior is intentional: the cascade trades
	efficiency for safety under uncertainty, reverting
	toward full expert coverage rather than producing
	confident misclassifications. Importantly, this
	conditional computation differs from deploying
	ExpertNet alone, which would incur full
	computational and operational cost across all inputs
	regardless of input quality. The cascade therefore
	acts as an adaptive routing mechanism, preserving
	baseline safety guarantees while enabling efficiency
	gains when input conditions permit.

	V11 fails differently under the same corruption.
	Rec-Healthy drops from 0.722 to 0.316 under sensor
	noise --- a 56\% reduction --- and F1-DGT drops from
	0.787 to 0.654 under field corruption
	(Table~\ref{tab:stress} and Figure~\ref{fig:stress}). These degradations occur
	simultaneously and without any compensating mechanism.
	V11 has no architectural isolation; corruption that
	enters the shared backbone \citep{howard2019} propagates
	to all task heads equally \citep{hendrycks2019,
		geirhos2019}. The cascade's advantage is not immunity
	to corruption --- the Sentinel's routing is clearly
	disrupted --- but that the disruption is contained to
	the routing decision rather than to the diagnostic
	output. Viewed more broadly, the cascade reframes
	classification as a form of risk allocation
	\citep{elkan2001, scott2005}, where uncertain cases
	are escalated rather than forced into potentially
	incorrect predictions.
	
	\subsection{Drought Recall as an Unresolved Problem}
	\label{sec:disc_drought}
	
	F1-DGT $= 0.791$ is the result that should give
	practitioners the most pause. The confusion matrix for
	Task 2 (Figure~\ref{fig:confusion}, center panel) shows
	the Expert correctly classifying 83\% of Drought cases
	and misclassifying 17\% as Weeds. In an insurance
	context where the agronomic response to drought stress
	--- irrigation, replanting, input substitution ---
	differs materially from the response to weed competition
	\citep{ceballos2019, barnett2007}, a 17\%
	misclassification rate on the minority damage class is
	not a minor modeling imperfection. It is a diagnostic
	reliability problem \citep{saleem2019, coulibaly2019}.
	
	FocalLoss with $\gamma = 2.0$ \citep{lin2017} partially
	addresses the 4:1 Weeds/Drought imbalance, and the
	Expert's convergence dynamics (Figure~\ref{fig:expert})
	show F1-WED rising steeply and stabilizing near 0.95
	by epoch 5 while F1-DGT plateaus near 0.79 after epoch
	5 and does not improve thereafter. This is not a training
	convergence failure \citep{loshchilov2019}. This
	suggests a degree of non-separability between drought
	and weed stress in RGB feature space at certain growth
	stages --- leaf yellowing, stunted growth, and canopy
	thinning occur in both conditions, particularly when
	canopy coverage is low and distinguishing features are
	less visually pronounced \citep{toda2019, sibiya2019}.
	While feature-space overlap is a plausible explanation,
	alternative factors such as label noise or
	stage-dependent annotation ambiguity in the mixed
	manual-automatic labeling pipeline \citep{waithaka2022,
		zhang2020} cannot be excluded without further analysis.
	Resolving this gap may require substantially more
	Drought-class examples \citep{chawla2002}, or auxiliary
	inputs --- multispectral bands, temporal sequences
	capturing damage progression \citep{coulibaly2019} ---
	that a single RGB smartphone image cannot provide
	\citep{waithaka2022, ceballos2019}.
	
	\subsection{Rec-Healthy at \texorpdfstring{$\tau =
			0.5$}{tau = 0.5}: A Feature, Not a Bug}
	\label{sec:disc_healthy}
	
	Rec-Healthy $= 0.424$ at $\tau = 0.5$
	(Figure~\ref{fig:confusion}, left panel) requires
	direct explanation rather than defensive qualification.
	The Sentinel was trained under a constraint that
	prioritizes $\text{Rec-Damaged} \geq 0.95$, with
	Rec-Healthy $\geq 0.40$ as a secondary floor
	\citep{elkan2001, cannon2002}. The checkpoint selected
	at epoch 4 --- Safety Score $= 0.8532$ (Figure~\ref{fig:sentinel}) --- satisfies both
	constraints. The consequence
	is a classifier that aggressively routes toward damage:
	57.6\% of Healthy crops are misrouted to ExpertNet as
	potentially damaged \citep{he2009, wallace2011}. This
	cost is not incidental --- it is the direct consequence
	of enforcing the Rec-Damaged constraint, and it was
	accepted by design \citep{elkan2001, scott2005}.
	
	The practical implication is that operators deploying
	at $\tau = 0.5$ should expect ExpertNet to process a
	substantial proportion of genuinely healthy crops
	\citep{ceballos2019}. At $\tau = 0.6$, Rec-Healthy
	improves to 0.650 --- still not high in absolute terms,
	but meaningfully better --- at the cost of Rec-Damaged
	falling to 0.929. Whether 0.929 is an acceptable
	safety floor is context-dependent \citep{barnett2007,
		barnett2008}. In a high-stakes insurance payout setting
	where every missed damage case has direct financial
	consequences for a vulnerable farmer \citep{jensen2016},
	it may not be. In a preliminary screening context where
	a human agronomist reviews ExpertNet outputs before any
	payout decision \citep{ceballos2019}, it may be
	entirely reasonable. For practitioners deploying in field conditions where
	expert review bandwidth is limited \citep{ceballos2019},
	$\tau = 0.6$ or $\tau = 0.7$ is likely the more
	realistic operating point: both deliver substantially
	higher Rec-Healthy (0.650 and 0.816 respectively)
	while maintaining Rec-Damaged above operationally
	meaningful thresholds \citep{barnett2007, jensen2016}.
	$\tau = 0.5$ should be reserved for contexts where
	the cost of any missed damage case is the dominant
	operational constraint.
	
	Taken together, the system's behavior is governed by
	the interaction between training constraints, threshold
	calibration, and input uncertainty \citep{minderer2021,
		ovadia2019}, rather than by model accuracy alone. This
	is the distinguishing property of a safety-constrained
	system: its deployment behavior is as much a function
	of its design objectives \citep{elkan2001, cannon2002}
	as of its learned representations \citep{howard2019,
		caruana1997}.
	
	\subsection{Toward Adaptive Routing}
	\label{sec:disc_adaptive}
	
	The fixed-threshold design of $\tau$ is a deliberate
	simplicity choice --- it requires no additional
	modeling, introduces no trainable parameters at
	deployment time, and is fully interpretable
	\citep{geifman2017}. But it is not the only possible
	design. Two extensions are natural.
	
	The first is uncertainty-aware routing, where the
	Sentinel's routing decision is based on predictive
	entropy rather than a fixed probability threshold. A
	sample with $P(\text{Damaged}|\mathbf{x}) = 0.55$ and
	high entropy --- indicating genuine model uncertainty
	\citep{ovadia2019} --- warrants Expert consultation
	more strongly than a sample with the same probability
	but low entropy. Fixed-threshold routing treats these
	identically; entropy-based routing would not
	\citep{geifman2017, wang2018skipnet}. This could reduce
	$\rho$ under clean conditions without sacrificing
	Rec-Damaged, by routing only genuinely ambiguous cases
	rather than all cases above a probability cutoff
	\citep{bolukbasi2017, figurnov2017}. However, the
	reliability of uncertainty estimates under distribution
	shift remains an open question \citep{ovadia2019,
		minderer2021} --- calibration may degrade precisely
	under the conditions where routing decisions become
	most critical \citep{taori2020}, inverting the intended
	behavior at the worst possible moment.
	
	The second is adaptive $\tau$ calibration, where $\tau$
	is adjusted dynamically based on observed batch
	statistics or domain shift indicators
	\citep{hendrycks2019, recht2019}. The threshold drift
	from $\tau = 0.56$ at epoch 4 to $\tau = 0.19$ at
	epoch 30 (Figure~\ref{fig:sentinel}, right panel) is evidence that optimal
	$\tau$ is sensitive to the model's confidence
	calibration \citep{minderer2021}. A deployment system
	that monitors routing rate $\rho$ and adjusts $\tau$
	to maintain a target $\rho$ could preserve efficiency
	under moderate corruption without retraining
	\citep{wang2018skipnet, huang2017}. Both extensions
	remain directions for future work; the current system
	provides the baseline against which they should be
	evaluated.

	
	\section{Limitations}
	\label{sec:limitations}
	
	The limitations of CascadeCropNet are not peripheral
	edge cases --- they are structural properties of the
	design that follow directly from the architectural
	choices and training objectives described in
	Sections~\ref{sec:method} and~\ref{sec:eval}. Stating
	them as such is more useful than treating them as
	caveats.
	
	The cascade introduces a hard error boundary at the
	triage stage. False negatives at the Sentinel ---
	truly damaged crops assigned
	$P(\text{Damaged}|\mathbf{x}) < \tau$ and therefore
	predicted Healthy --- never reach ExpertNet. In a flat
	classifier, errors are not gated by an irreversible
	intermediate decision, and all samples remain available
	to downstream representations \citep{caruana1997,
		standley2020}. The cascade localizes failure at the
	routing stage, making Sentinel false negatives both
	identifiable in structure and lost to the system within
	the current routing configuration \citep{geifman2017}.
	No downstream correction mechanism exists within the
	pipeline. At $\tau = 0.5$, this means 2.6\% of truly
	damaged crops are missed by the system entirely
	(Table~\ref{tab:sweep}) --- a figure that should be
	understood not as an acceptable error rate but as a
	hard floor imposed by the Sentinel's recall at this
	operating point \citep{elkan2001, cannon2002}. In an
	insurance-linked context \citep{barnett2007, jensen2016},
	each such case represents a farmer who sustained real
	losses and received no payout.
	
	Failure in the cascade is asymmetric in a deeper sense
	than the false negative rate alone captures. 
	 While not
	evaluated empirically in this work, under more severe
	distribution shift affecting Sentinel recall ---
	geographic transfer to regions with substantially
	different damage presentations \citep{ramcharan2017,
		picon2019}, or sensor degradation beyond the corruption
	levels tested here \citep{hendrycks2019} --- the
	cascade may underperform flat models entirely, as
	Sentinel false negatives accumulate faster than the
	architectural isolation property can compensate
	\citep{taori2020, recht2019}. More precisely: because
	routing decisions are irreversible, degradation in
	Sentinel recall disproportionately impacts system
	performance, and improvements in downstream Expert
	accuracy cannot compensate for degraded Sentinel recall
	\citep{geifman2017, scott2005}.
	
	Drought classification remains an open problem within
	the current system. F1-DGT $= 0.791$ is insufficient
	for treatment-level diagnostic reliability in contexts
	where the response to drought stress differs materially
	from the response to weed competition \citep{ceballos2019,
		barnett2007}. The 4:1 Weeds/Drought class imbalance
	\citep{he2009, chawla2002} is a primary driver, though
	feature-space overlap between drought and weed stress
	in RGB feature space at certain growth stages
	\citep{toda2019, sibiya2019}, and potential label noise
	or stage-dependent annotation ambiguity in the mixed
	manual-automatic labeling pipeline \citep{waithaka2022,
		zhang2020}, likely contribute as well. FocalLoss with
	$\gamma = 2.0$ \citep{lin2017} partially addresses the
	imbalance but does not resolve the underlying
	non-separability \citep{cui2019}. Per-class
	oversampling \citep{chawla2002}, synthetic augmentation
	targeting Drought presentations \citep{shorten2019},
	or auxiliary inputs beyond RGB \citep{coulibaly2019}
	are candidate directions; none has been evaluated in
	this work.
	
	Expert performance is reported on truly damaged samples
	under the intrinsic evaluation protocol described in
	Section~\ref{sec:eval}. Experiment~6
	(Section~\ref{sec:res_hypotheses}) quantifies the
	consequence of this choice. The intrinsic--TP-routed
	protocol gap is $-0.0004$ at $\tau = 0.5$, confirming
	that the intrinsic protocol accurately reflects
	ExpertNet's performance on correctly-routed samples.
	However, the end-to-end 3-class F1-macro is 0.767
	against the intrinsic F1-Expert of 0.932 --- a gap
	of 0.165 that arises entirely from false positive
	routing. At $\tau = 0.5$, 55\% of healthy crops are
	routed to ExpertNet and receive a false damage
	diagnosis; 90\% of these false diagnoses are WED
	(weed presence), reflecting a systematic visual
	similarity between healthy crops and the weed-affected
	class. Practitioners should monitor end-to-end
	3-class accuracy, not intrinsic Expert F1, when
	evaluating deployed system performance.

	A structural training asymmetry also warrants
	acknowledgement. V11 is trained on all three classes
	(Healthy, DGT, WED) using the full training set;
	ExpertNet is trained only on the damaged subset
	(Section~\ref{sec:masking}).
	This is not purely an architectural advantage --- it
	is a form of curriculum learning and conditional
	filtering that may improve performance independently
	of the cascade structure. Experiment~2 isolates the
	representational gap to $+0.008$ F1-macro under fair
	evaluation, but does not fully disentangle training
	distribution from architecture. A fully controlled
	comparison would require training a flat model on the
	damaged subset only, which has not been conducted.
	
	The evaluation is conducted on a single dataset from a
	specific geographic and agronomic context ---
	smallholder maize farms in eight Kenyan counties during
	2020--2021. The cascade architecture and the
	$\tau$-based control mechanism are general, but the
	specific performance values reported
	($\text{Rec-Damaged} = 0.974$, F1-Expert $= 0.868$,
	Safety Score $= 0.8532$) reflect this context.
	Transfer to other crops, geographies, or smartphone
	camera distributions would require local revalidation
	of $\tau$ and potentially retraining of both models.
	F1-DGT $= 0.791$ specifically limits treatment-level
	deployment for drought diagnosis --- this is a
	structural limitation of the diagnostic stage under
	the current class imbalance and RGB feature space, not
	of the cascade architecture itself.

	The system's operational behavior depends critically on
	the calibration of Sentinel output probabilities and
	the selection of $\tau$ on validation data
	\citep{minderer2021, ovadia2019}. Under deployment
	conditions where probability calibration shifts --- due
	to geographic transfer \citep{ramcharan2017}, sensor
	change \citep{hendrycks2019}, or distributional drift
	over time \citep{recht2019} --- the operating
	characteristics associated with a fixed $\tau$ may no
	longer hold. A $\tau$ selected to achieve
	$\text{Rec-Damaged} = 0.974$ on the Kenyan validation
	set \citep{waithaka2022} may produce substantially
	different recall under a different deployment
	distribution \citep{taori2020}. Periodic revalidation
	of $\tau$ on locally representative data is advisable
	\citep{ceballos2019, barnett2008}, and the
	$\tau$-as-Pareto-parameter framing in
	Section~\ref{sec:rationale} is designed precisely to
	make this recalibration tractable without retraining.
	This does not eliminate the dependency on local
	calibration data \citep{minderer2021}.

	While the optimal threshold $\tau$ may vary under
	distribution shift and requires local revalidation,
	the trade-off structure between Rec-Healthy and
	Rec-Damaged remains monotonic across all evaluated
	settings (Table~\ref{tab:sweep}). This ensures
	that threshold adjustment is
	predictable and preserves the intended
	safety--efficiency trade-off, even when deployment
	conditions differ from the training distribution.

	A methodological asymmetry in the evaluation design
	warrants explicit acknowledgement as a structural
	limitation rather than a footnote. The threshold sweep
	--- from which the two primary operating points
	($\tau = 0.5$, $\tau = 0.6$) are derived --- is
	conducted on the validation set, while the robustness
	stress test is conducted on the held-out test set.
 This means the paper's
	two central results are not directly comparable as
	absolute numbers: the safety performance figures
	(Section~\ref{sec:res_safety}) and the robustness
	figures (Section~\ref{sec:res_robust}) come from
	different data partitions. Harmonizing both evaluations
	to a single held-out split --- by running either the
	threshold sweep on the test set or the stress test on
	the validation set --- would resolve this inconsistency
	and should be prioritized in follow-on work
	\citep{menardi2014}.
	
	More broadly, the architecture imposes a structural
	coupling between safety and efficiency: the recall
	constraint that enforces the safety floor simultaneously
	drives aggressive routing \citep{elkan2001, cannon2002},
	and any improvement in Rec-Damaged at a given $\tau$
	comes at the cost of increased expert load $\rho$
	\citep{geifman2017, bolukbasi2017}. This is not a
	property that can be engineered away within the current
	design --- it is a consequence of the constrained
	optimization formulation in
	Section~\ref{sec:rationale} \citep{scott2005}.
	Relaxing the coupling would require either a more
	expressive Sentinel backbone \citep{tan2019, he2016}
	that achieves better class separation, or a routing
	mechanism that conditions on richer signals than scalar
	damage probability alone \citep{wang2018skipnet,
		figurnov2017}.

	The cascade architecture represents one concrete
	instantiation of enforcing safety constraints
	through structured inference. Alternative
	approaches, such as multi-head or hierarchical
	loss formulations, may achieve similar predictive
	performance but do not provide the same structural
	separation of triage from diagnosis at inference
	time. In contrast,
	the cascade enforces safety through conditional
	execution, ensuring that uncertain or high-risk
	cases are systematically escalated. Empirical
	comparison with such alternatives is left for
	future work.

   The cascade provides structural
   interpretability: the decision pathway is
   explicit and traceable, with triage and
   diagnosis separated into distinct,
   inspectable stages. It does not provide
   epistemic interpretability. The system does
   not quantify predictive uncertainty or
   indicate when its outputs should be
   distrusted, which limits its use in
   risk-sensitive decision workflows such as
   agricultural insurance where operators need
   to know not only \emph{where} a decision
   occurred but \emph{how confident} the system
   was when it made it. Deterministic routing
   policies of this form make failure
   analyzable but not detectable at inference
   time, as the system lacks mechanisms to
   signal when its predictions may be
   unreliable. Uncertainty-aware routing,
   discussed in Section~\ref{sec:disc_adaptive},
   is one candidate direction toward closing
   this gap.

	
	\section{Conclusion}
	\label{sec:conclusion}
	
	This work addresses a fundamental mismatch between the
	asymmetric cost structure of insurance-linked crop
	damage assessment and the objectives typically used to
	train automated classification systems
	\citep{elkan2001, cannon2002, ceballos2019}. Prior work
	has formalized asymmetric risk through cost-sensitive
	learning \citep{elkan2001, zadrozny2003} and
	constrained classification \citep{cannon2002, scott2005}
	--- the theoretical apparatus exists. What has received
	less attention is how these ideas translate into systems
	that maintain safety-aligned behavior when input
	distributions shift at deployment \citep{hendrycks2019,
		taori2020, ovadia2019}. This paper proposes an
	architectural response: a hierarchical cascade that
	explicitly separates binary triage from fine-grained
	diagnosis, enforces a recall constraint at the level of
	model selection during training, and structurally
	isolates diagnostic inference from corruption affecting
	triage inputs.
	
	Empirically, the cascade exhibits strong robustness
	under controlled corruption applied to Sentinel inputs.
	Across all tested conditions, ExpertNet maintains
	stable diagnostic performance --- F1-Expert $= 0.877$,
	F1-DGT $= 0.805$, F1-WED $= 0.949$ --- while the flat
	baseline degrades by 17 to 35 percent across the same
	metrics under the same conditions
	(Table~\ref{tab:stress}). These results suggest that
	the observed robustness is closely linked to the
	architectural isolation between stages, which
	constrains how perturbations propagate through the
	system \citep{hendrycks2019, geirhos2019}. This
	interpretation is based on the corruption regimes
	evaluated in this study and does not establish
	invariance under broader forms of distribution shift
	\citep{recht2019, taori2020}, including geographic
	transfer beyond the Kenyan training distribution
	\citep{waithaka2022} or sensor configurations not
	represented in the test set \citep{ceballos2019}.
	
	The routing threshold $\tau$ provides a practical
	mechanism to adapt system behavior at deployment
	without retraining \citep{geifman2017, bolukbasi2017}.
	By adjusting $\tau$, operators can navigate the
	trade-off between expert workload and recall of damaged
	crops along a Pareto frontier whose axes are directly
	interpretable: expert load $\rho$ and Rec-Damaged
	\citep{scott2005, cannon2002}. At $\tau = 0.5$,
	$\text{Rec-Damaged} = 0.974$ with a 13.2\% reduction
	in expert inference volume; at $\tau = 0.6$,
	$\text{Rec-Damaged} = 0.929$ with a 22.6\% reduction
	(Table~\ref{tab:sweep}). Under corruption, the
	system exhibits a conservative failure mode --- routing
	becomes more aggressive, increasing expert load while
	preserving diagnostic quality \citep{ovadia2019,
		minderer2021} --- consistent with the safety
	requirements of workflows where missed damage carries
	greater operational cost than unnecessary review
	\citep{barnett2007, jensen2016, ceballos2019}.
	
	
	Several limitations remain and should be understood
	as structural rather than incidental. Drought
	classification at F1-DGT $= 0.791$ is insufficient
	for reliable treatment-level diagnosis, likely
	reflecting a combination of class imbalance
	\citep{he2009, chawla2002}, visual overlap between
	drought and weed stress in RGB feature space
	\citep{toda2019, sibiya2019}, and potential annotation
	noise in the mixed labeling pipeline
	\citep{waithaka2022, zhang2020}. The mapping between
	$\tau$ and system recall depends on Sentinel
	probability calibration \citep{minderer2021}, which
	may shift under geographic transfer
	\citep{ramcharan2017} or changing sensor conditions
	\citep{hendrycks2019}; deployment in new regions
	requires local revalidation to maintain the intended
	safety guarantees \citep{barnett2008, ceballos2019}.
	The intrinsic-deployed performance gap for ExpertNet
	--- evaluated here on truly damaged samples rather
	than sentinel-routed samples
	(Section~\ref{sec:eval}) --- remains empirically
	unquantified and constitutes an open problem for
	follow-on work \citep{ovadia2019, davis2006}.
	
	More broadly, this work suggests that incorporating
	decision structure at the architectural level may
	offer a more reliable pathway to handling asymmetric
	risk under distribution shift than loss-based
	approaches alone \citep{elkan2001, cannon2002,
		scott2005}. The cascade is one concrete instantiation
	of this idea, grounded in the conditional label
	structure of the Eyes on the Ground dataset
	\citep{waithaka2022} and the operational requirements
	of picture-based agricultural insurance
	\citep{ceballos2019, barnett2007, jensen2016}.
	Whether it generalizes across domains and task
	structures where similar cost asymmetries and
	distribution shift conditions hold
	\citep{hendrycks2019, taori2020} is a question that
	the results here motivate but do not answer.
	
	\section*{CRediT authorship contribution
		statement}
	\textbf{Jos\'{e} T.~M. Hagbe:}
	Conceptualization, Methodology, Software,
	Formal analysis, Investigation, Data curation,
	Writing --- original draft, Writing --- review
	\& editing, Visualization.
	\textbf{Kawsar Gounou:}
	Writing --- review \& editing, Validation.
	\textbf{Songbian Zim\'{e}:}
	Writing --- review \& editing, Validation.
	
	\section*{Declaration of competing interest}
	The authors declare that they have no known
	competing financial interests or personal
	relationships that could have appeared to
	influence the work reported in this paper.
	
	\section*{Funding}
	This research received no specific grant from
	any funding agency in the public, commercial,
	or not-for-profit sectors.
	
	\section*{Ethical statement}
	This study uses the publicly available Eyes on
	the Ground dataset \citep{waithaka2022},
	collected under informed consent as part of a
	picture-based insurance protocol
	\citep{ceballos2019}. No new human or animal
	subjects were involved in this research.
	
	\section*{Data availability}
	The Eyes on the Ground dataset is publicly
	available on Radiant MLHub at
	\url{https://doi.org/10.34911/rdnt.1bs2jw}
	\citep{waithaka2022} under a CC-BY-SA-4.0
	license. Model checkpoints and evaluation
	scripts are available from the corresponding
	author upon reasonable request.
	
	\section*{Acknowledgements}
	The Eyes on the Ground dataset was produced
	by ACRE Africa, the International Food Policy
	Research Institute, the Lacuna Fund, KALRO,
	BlueGreen Labs, and Dvara E-Registry
	\citep{waithaka2022}. The authors thank the
	smallholder farmers who contributed images
	under the picture-based insurance protocol
	\citep{ceballos2019}.

	\linenumbers

\bibliography{refs}

\end{document}